\documentclass[a4paper,conference]{IEEEtran}
\ifCLASSINFOpdf
\else
\fi

\usepackage{times}
\usepackage{soul}
\usepackage{url}
\usepackage[hidelinks]{hyperref}
\usepackage[utf8]{inputenc}
\usepackage[small]{caption}
\usepackage{graphicx}
\usepackage{amsmath}
\usepackage{amsthm}
\usepackage{booktabs}
\usepackage{algorithm}
\usepackage{algorithmic}
\usepackage{newfloat}
\usepackage{listings}
\usepackage{amsmath}
\usepackage{amssymb}
\usepackage{bm}
\usepackage{booktabs}
\usepackage{colortbl}
\usepackage{siunitx}
\usepackage{lipsum}
\usepackage{multirow}
\usepackage{tabularx}
\usepackage{adjustbox}
\usepackage{tikz}
\usepackage{xspace}
\usepackage{bbm}
\usetikzlibrary{bayesnet}
\usepackage{wrapfig}
\usepackage{subcaption}
\usepackage{dsfont}

\newcolumntype{L}{>{\raggedright\arraybackslash}X}
\floatstyle{ruled}
\newfloat{listing}{tb}{lst}{}
\floatname{listing}{Listing}

\newcommand{\daren}{\texttt{DAReN}\xspace}

\renewcommand*{\figureautorefname}{Fig.}
\renewcommand*{\tableautorefname}{Tab.}
\renewcommand*{\sectionautorefname}{Sec.}

\renewcommand*{\equationautorefname}{Eq.}

\begin{document}
%
\title{DAReN: {A} {C}ollaborative {A}pproach {T}owards Visual {R}easoning {A}nd {D}isentangling}


\author{\IEEEauthorblockN{Pritish Sahu, Kalliopi Basioti, Vladimir Pavlovic}
\IEEEauthorblockA{Department of Computer Science, Rutgers University}
{\{pritish.sahu, kalliopi.basioti\}@rutgers.edu, vladimir@cs.rutgers.edu}
}

\maketitle

\definecolor{redcol}{rgb}{1, 0, 0}
\definecolor{bluecol}{rgb}{0, 0, 1}
\newcommand{\red}[1]{\textcolor{redcol}{#1}}
\newcommand{\blue}[1]{\textcolor{bluecol}{#1}}
\renewcommand{\paragraph}[1]{\smallskip\noindent{\bf{#1}}}

\newcommand{\todo}[1]{\red{TODO: {#1}}}
\newcommand{\colons}[1]{``{#1}''}
\newcommand{\tb}[1]{\textbf{#1}}
\newcommand{\mb}[1]{\mathbf{#1}}
\newcommand{\bs}[1]{\boldsymbol{#1}}
\def\ith#1{#1^\textit{th}}

\renewcommand{\vec}[1]{\mathbold{#1}}
\newcommand{\mat}[1]{\mathbold{#1}}
\newcommand{\vx}{\vec{x}}
\newcommand{\vX}{\mat{X}}

\newcommand{\secref}[1]{Section~\ref{sec:#1}}
\newcommand{\figref}[1]{Figure~\ref{fig:#1}}
\newcommand{\tabref}[1]{Table~\ref{tab:#1}}
\newcommand{\eqnref}[1]{\eqref{eq:#1}}
\newcommand\RotText[1]{\rotatebox[origin=c]{90}{\parbox{1cm}{\centering#1}}}

\def\algorithmautorefname{Algorithm}
\def\figureautorefname{Figure}
\def\tableautorefname{Table}
\def\equationautorefname{Eq.}
\def\sectionautorefname{Section}
\def\etal{et~al.\_} 
\def\eg{e.g.,~} 
\def\ie{i.e.,~} 
\def\etc{etc} 
\def\cf{cf.~} 
\def\viz{viz.~} 
\def\vs{vs.~} 
\def\newtext#1{\textcolor{blue}{#1}}
\def\modtext#1{\textcolor{red}{#1}}

\ifx \compilemain \undefined

\begin{abstract}
Computational learning approaches to solving visual reasoning tests, such as Raven's Progressive Matrices (RPM), critically depend on the ability to identify the visual concepts used in the test (i.e., the representation) as well as the latent rules based on those concepts (i.e., the reasoning).  However, learning of representation and reasoning is a challenging and ill-posed task, often approached in a stage-wise manner (first representation, then reasoning).
In this work, we propose an end-to-end joint representation-reasoning learning framework, which leverages a weak form of inductive bias to improve both tasks together.  Specifically, we introduce a general generative graphical model for RPMs, \textit{GM-RPM}, and apply it to solve the reasoning test. We accomplish this using a novel learning framework \tb{D}isentangling based \tb{A}bstract \tb{R}easoning \tb{N}etwork (\daren) based on the principles of  \textit{GM-RPM}.
We perform an empirical evaluation of \daren over several benchmark datasets.
\daren shows consistent improvement over state-of-the-art (SOTA) models on both the reasoning and the disentanglement tasks.
This demonstrates the strong correlation between disentangled latent representation and the ability to solve abstract visual reasoning tasks.
\end{abstract}

\section{\tb{Introduction}}
\label{sec:intro}

Raven's Progressive Matrices (RPM)~\cite{carpenter1990one,raven1941standardization,raven1998raven} is a widely acknowledged metric in the research community to test the cognitive skills of humans.
RPM is primarily used to assess lateral thinking, i.e., the ability to systematically process the results and find solutions to unseen problems without drawing on prior knowledge~\cite{klein2018scrambled}.
The vision community has often employed Raven's test to evaluate the abstract reasoning skills of an AI model~\cite{hoshen2017iq,little2012bayesian,lovett2017modeling,lovett2010structure,lovett2009solving}.
\figref{demo} illustrates a RPM question, given a $3 \times 3$ matrix, where each cell contains a visual geometric design except for the last cell in the bottom row. An AI model must pick the best-fit image from a list of six to eight choices to complete the matrix. Solving the question requires figuring out the underlying rules in the matrix as shown in \figref{demo} where the correct rule is ``color is constant in a row'', which eliminates other choices leaving choice ``A'' as the correct answer.
An AI model needs to infer the underlying rules in the top two rows or columns to fill the missing piece in the last row.
The visual IQ score of the above model obtained via solving abstract reasoning tasks can provide ground to compare AI against human intelligence.

\begin{figure}[!ht]
    \begin{center}
        \includegraphics[width=0.9\linewidth]{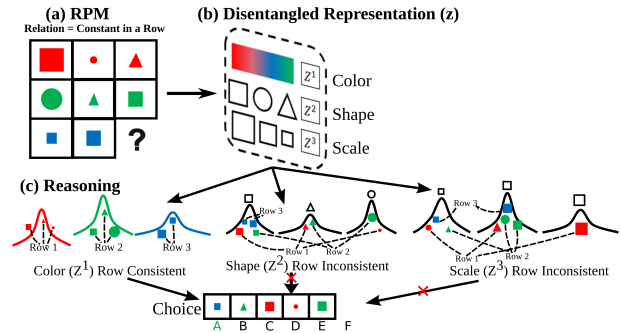}
    \end{center}
    \caption{\tb{a:} RPM instance with rule ``constant-in-a-row''. \tb{b}. The true generative factors in the data. \tb{c}. \tb{Reasoning with factors:} The distribution of $z^{i}$ is represented by a Gaussian. Only $z^1$ (color) satisfies the rule while the other two factors are inconsistent. \tb{Index one} in the choice list is the correct answer.}
    \label{fig:demo}
\end{figure}

Earlier computation models depended on handcrafted heuristics rules on propositions formed from visual inputs to solve RPM~\cite{bringsjord2003artificial,lovett2007analogy}.
The lack of success in the previous approaches and the inclusion of large RPM datasets~\cite{van2019disentangled}, PGM~\cite{barrett2018measuring}, RAVEN~\cite{zhang2019raven} facilitated the employment of neural networks to solve abstract reasoning tasks~\cite{hoshen2017iq,barrett2018measuring,zhang2019raven,van2019disentangled,steenbrugge2018improving}.
Until now, these abstract reasoning methods have employed existing deep learning methods such as CNN~\cite{lecun1989backpropagation},ResNet~\cite{he2016deep} to improve reasoning but largely ignore to learn the visual attributes as independent components.
Even though these models have improved abstract reasoning tasks, the performance is still sub-optimal compared to humans.
These setbacks to the model performance are caused due to the lack of adequate and task-appropriate visual representation.
The model should learn to separate the key attributes needed for reasoning as independent components.

These critical attributes, aka disentangled representations, \cite{bengio2013representation,ridgeway2016survey} break down the visual features to their independent generative factors, capable of generating the full spectrum of variations in the ambient data.
We argue that a better-disentangled model is essential for the better reasoning ability of machines.
A recent study \cite{locatello2019challenging} via impossibility theorem has shown the limitation of learning disentanglement independently.
The impossibility theorem states that without any form of inductive bias, learning disentangled factors is impossible.
Since collecting label information of the generative factors is challenging and almost impossible in real-world datasets, previous works have focused on some form of semi-supervised or weakly-supervised methods. Few of the prior works in  disentanglement using inductive bias involve ~\cite{kingma2014semi} that uses a subset of ground truth factors,  ~\cite{bouchacourt2018multi} that formed pair of images with common visual attributes on a subset of factors, and   ~\cite{wang2014learning} where the factors in a pair of images are ranked  on the subset of factors.   
Our work improves upon the model's reasoning ability by using the inductive reasoning present in the spatial features. 
Utilizing the underlying reasoning, i.e., rules on visual attributes in RPM induces weak supervision that helps improve disentanglement, leading to better reasoning.



\cite{van2019disentangled} investigated the dependency between ground truth factor of variations and reasoning performance. We take a step further and consider jointly learning disentangled representation and learning to reason (critical thinking). Unlike the above-proposed model, i.e. (working in a staged process to improve disentangling or improve downstream accuracy),  we work on the weakness of both components and propose a novel way to optimize both in a single end-to-end trained model. We demonstrate the benefits of the interaction between representation learning and reasoning ability. Our motivation behind using the same evaluation procedure by \cite{van2019disentangled} is as follows: 1) the strong visual presence, 2) information of the generative factors help in demonstrating the model efficacy on both reasoning accuracy and disentanglement (strong correlation), 3) possibility of comparing the disentangled results with state-of-the-art disentangling results.

In summary, the contributions of our work are threefold:
(1) We  propose a general generative graphical model for RPM, GM-RPM, which will form the essential basis for inductive bias in joint learning for representation + reasoning.
(2) Building upon GM-RPM, we propose a novel learning framework named \textbf{D}isentangling based \textbf{A}bstract \textbf{Re}asoning \textbf{N}etwork (\daren) composed of two primary components -- disentanglement network, and reasoning network. It learns to disentangle factors and uses the representation to detect the underlying relationship and the object property used for the relation. To our knowledge, (\daren) is the firs joint learning framework that separates the underlying generative factors and solve reasoning.
(3) We show that \daren outperforms all state-of-the-art baseline models in reasoning and disentanglement metrics, demonstrating that reasoning and disentangled representations are tightly related; learning both in unison can effectively improve the downstream reasoning task.


\section{\tb{Related Works}}
\label{sec:rel_works}

\textbf{Visual Reasoning.} \quad
Solving RPM have recently gained much attention due to their high correlation with human intelligence~\cite{raven1998raven,barrett2018measuring,zhang2019raven}.
Initial works based on rule-based heuristics such as symbolic representations ~\cite{carpenter1990one,lovett2017modeling,lovett2010structure,lovett2009solving} or relational structures ~\cite{little2012bayesian,mcgreggor2014confident,mekik2018similarity} failed to comprehend the reasoning tasks due to their underlying assumptions.
These assumptions include access to the symbolic representations of images, domain expertise on the underlying operations, and comparisons that help solve the task.
~\cite{wang2015automatic} proposed a systematic way of automatically generating RPM using first-order logic to try to understand and solve these tasks fully.
Growing interest introduced two RPM dataset ~\cite{barrett2018measuring,zhang2019raven}, which led to significant progress in solving reasoning tasks~\cite{hu2020hierarchical,benny2021scale, hoshen2017iq,barrett2018measuring}.

\textbf{Disentanglement.} \quad
Recovering independent data generating ground truth factors is a well-studied problem in machine learning.
In recent years there is renewed interest in unsupervised learning of disentangled representations~\cite{higgins2016beta,kim2018disentangling,kim2019bayes,kim2019relevance,locatello2019challenging,ridgeway2016survey,tschannen2018recent}.
Nevertheless, this research area has not reached a major consensus on two major notions: i) no widely accepted formalized definition~\cite{bengio2013representation,ridgeway2016survey,locatello2019challenging,tschannen2018recent}, ii) no single robust evaluations metrics to compare the models~\cite{burgess2018understanding,kim2018disentangling,chenisolating,eastwood2018framework,kumar2017variational}.
However, the key fact common in all models is the recovery of statistically independent~\cite{ridgeway2016survey} learned factors.
A majority of the research follows the definition presented in~\cite{bengio2013representation}, which states that the underlying generative factors correspond to independent latent dimensions, such that changing a single factor of variation should change only a single latent dimension while remaining invariant to others.
Recent work~\cite{locatello2019challenging}  showed that it is impossible to learn disentangled representation without the presence of inductive bias, prompting the shift to semi-supervised~\cite{locatello2019disentangling,sorrenson2020disentanglement,khemakhem2020variational} and weak-supervised~\cite{locatello2020weakly,bouchacourt2018multi,hosoya2018group,shu2019weakly} disentangling models.

Recent reasoning works \cite{hu2020hierarchical,benny2021scale} have focused on Raven \cite{zhang2019raven}, PGM \cite{barrett2018measuring} datasets for evaluation.
In contrast to the above, we focus on learning both disentangled representation and solving abstract visual reasoning.
Our work is inspired by the large-scale study in~\cite{van2019disentangled}, suggesting dependence between learning disentangled representations and solving visual reasoning.
Our proposed framework leverages join learning, which improves both the reasoning and the disentanglement performance.
Since we quantify disentanglement score along with reasoning accuracy, the datasets used in ~\cite{van2019disentangled} are well suited compared to Raven~\cite{zhang2019raven}, PGM~\cite{barrett2018measuring} which are not adapted for quantitative evaluation of disentanglement.

\section{\tb{Problem Formulation and Approach}}
\label{sec:approach}
We begin by describing the problem of RPM in the domain of visual reasoning task in \secref{problemstatement}, where we elaborate on the process of what constitutes valid RPM.
Next, we propose our general generative graphical model for RPM, \textit{GM-RPM}, which will form the essential basis for inductive bias in joint learning for representation and reasoning in \secref{gmrpm}.
Finally, in  \secref{daren}, we describe our learning framework a.k.a \textbf{D}isentangling based \textbf{A}bstract \textbf{R}easoning \textbf{N}etwork (\daren) based on a variational autoencoder (VAE) and a reasoning network for joint representation-reasoning learning.

\subsection{\tb{Visual Reasoning Task}}
\label{sec:problemstatement}
The Raven's matrix denoted as $\mathcal{M}$, of size $M \times M$ contains images at all $i,j$ location except at $\mathcal{M}_{MM}$. 
The aim is to find the best fit image $a^*$ at  $\mathcal{M}_{MM}$ from a list of choices denoted as $A$.
For our current work, we follow the procedure by \cite{van2019disentangled} to prepare RPM.
Similar to prior work, we have fixed $M = 3$, where \( \mathcal{M} = \{x_{11}, \ldots, x_{32}\} \) in row-major order and $\mathcal{M}_{33}$ is empty that needs to be placed with the correct image from the choices.
We also set the number of choices $|A|= 6 $, where \( A = \{a_1, \ldots, a_6\} \).
We improve upon the prior work by formulating an abstract representation for the matrices $\mathcal{M}$ by defining a structure $S$ on the image attributes ($o$) and relation types ($r$) applied to the image attributes:
\[ S = \{ (r, o) : r \in R  \text{ and } o \in O \}.\]
The set $R$ consists of relations proposed by \cite{carpenter1990one} that are constant in a row, quantitative pairwise progression, figure addition or subtraction, distribution of three values, and distribution of two values.
We assume images are generated from underlying ground truth data generative factors ($K$) that constitute RPM.
These image factors ($O$) consist of the object type, size, position (XY--axis), and color.
The structure $S$ is a set of tuples, where each tuple is formed by randomly sampling a relation from $R$ and image attribute from $O$.
For instance, if $S$ = \{(constant in a row, color), (quantitative pairwise progression, size)\}, every image in each row of $\mathcal{M}$  will have the same (constant) value for attribute color, and progression relation instantiated on size of images from left to right.
This set of $S$ can contain a max of $|R| * |O|$ tuple, where the problem difficulty rises with the increase in the size of $R$, $O$ or $|S|$ or any combination of them.


\paragraph{Generating RPM.}
Using $S$, multiple realizations of the matrix $\mathcal{M}$ are possible depending on the randomly sampled values of $(r,o)$.
We use $\bm{o}$ to denote the image attributes in $S$, in the example above  $S= \{ \text{color}, \text{object type} \}$ \footnote{We interchangeably use $\bm{o}$ to denote the subset of image attributes that adhere to rules of RPM as well as the multi-hot vector $\bm{o} \in \{0,1\}^K$ whose non-zero values index those attributes.}.
In the generation process, we sample values for attributes in $\bm{o}$ that adhere to their associated relation $r$ and the values for image attributes in $\overline{\bs{o}} = O \setminus o$, that are not part of $S$, are sampled randomly for every image.
Next, we sample images at every $\mathcal{M}_{ij}$ where the image attribute values matches with the values sampled above for $\bm{o} \cup \overline{\bs{o}}$.
For the matrix $\mathcal{M}$ to be a valid RPM the sampled values for $\overline{\bs{o}}$ must not comply with the relation set $r \in S$ across all rows in $\mathcal{M}$.
However, in the above example where $\overline{\bs{o}} = \{$position, size$\}$, a valid $\mathcal{M}$ can also have the same values for position or size (or both) in a row as long as they do not adhere to any relations in $S$ for more than one row.
The above is an example of a distractor, where the attributes in $\overline{\bs{o}}$ during the sampling process might satisfy some $(r,o) \in S$ for any one row in $\mathcal{M}$ but not for all $M$ rows.
These randomly varying values in $\overline{\bs{o}}$ add a layer of difficulty towards solving RPM.
The result of the above generation steps produces a valid RPM matrix $\mathcal{M}$.
The task of any model trained to solve $\mathcal{M}$ has to find $r$ that is consistent across all rows or columns in $\bm{o}$ and discard the distracting features $\overline{\bm{o}}$.
In the rest of the paper, we focus on the row-based relationship in RPM\footnote{Our solution could trivially be extended to address columns or both rows and columns.}.


\subsection{\tb{Inductive Prior for RPM (GM-RPM)}}
\label{sec:gmrpm}
While previous works have made strides in solving RPM \cite{little2012bayesian,lovett2017modeling,van2019disentangled}, the gap in reasoning and representation learning between those approaches and the human performance remains. To narrow this gap, we propose a minimal inductive bias in the form of a probabilistic graphical model described here that can be used to guide the joint representation-reasoning learning.
\begin{figure}[!ht]
    \vspace{0.2cm}
    \begin{center}
        \resizebox{0.75\columnwidth}{!}{
            \tikzstyle{dlatent} = [rectangle,fill=white,draw=black,inner sep=1pt,
minimum size=20pt, font=\fontsize{10}{10}\selectfont, node distance=1]












\begin{tikzpicture}

    \node[dlatent, yshift=0.5cm] (rule) {$r$} ; %
    
    \node[dlatent] at(0.0,-0.3) (att) {$\mb{o}$} ; %

    \node[latent, right=of rule] (zo) {$\mb{z}_{i:,o}$} ; %
    \node[latent, below=of zo] (zoc) {$\mb{z}_{ij,\overline{o}}$} ; %

    \node[dlatent, above=of zo] (okn) {$\mb{o}_{KN}$} ; %

    \node[latent, right=of zoc] (z) {$\mb{z}_{ij}$} ; %

    \node[latent, below=of zoc, yshift=0.5cm] (zn) {$\mb{z}_{ij,n}$} ; %

    \node[obs, right=of z] (x) {$\mb{x}_{ij}$} ; %

    \plate[inner sep=0.25cm, xshift=-0.12cm, yshift=0.12cm] {plate0} {(zoc) (zn) (x)} {columns: $M$}; %

    \plate[inner sep=0.25cm, xshift=-0.12cm, yshift=0.12cm] {plate1} {(zo) (plate0)} {rows: $M$}; %
    \plate[inner sep=0.25cm, xshift=-0.12cm, yshift=0.12cm] {plate2} {(rule) (att) (plate1)} {RPM: $D$}; %

    \edge {rule} {zo} ; %
    \edge {zo} {z} ; %
    \edge {zoc} {z} ; %
    \edge {zn} {z}; %
    \edge {z} {x} ; %
    \edge {okn} {z} ; %
    \edge {att} {z} ; %

\end{tikzpicture}
        }
    \end{center}
    \caption{Generative model for RPM. See
    \autoref{sec:gmrpm} for details. }
    \label{fig:gmrpm}
\end{figure}
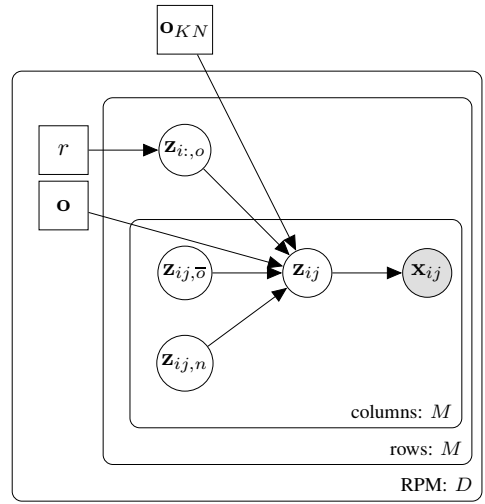
\figref{gmrpm} defines the structure of the general generative graphical model for RPM.  This model describes an RPM $\mathcal{M} = \{ \bs{x}_{11}, \ldots, \bs{x}_{MM} \}$, where $\mb{x}_{ij}, i,j= 1, \ldots, M $, denote the images in the puzzle, with the correct answer at $x_{MM}$, defined by rule $r$ and factors $o$.



Latent vectors $\bm{z}_{ij} \in \Re^{K+N}$ are the representations of the $K$ attributes, to be learned by our approach, and some inherent noise process encompassed in the remaining $N$ dimensions of $\bm{z}_{ij}$, $\bm{z}_{ij,n} \in \Re^N$, which we refer to as nuisances.
\begin{wrapfigure}{r}{0.5\linewidth}
  \begin{center}
    \includegraphics[width=0.98\linewidth]{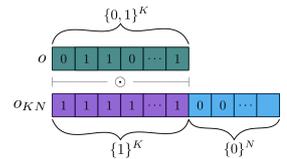}
  \end{center}
  \caption{Illustration of the indexing vectors $\mb{o}$ and $\mb{o}_{KN}$. `1' entries in $\mb{o}$ indicate the relevant factors used in a puzzle instance.  $N$ `0' entries (blue) in $\mb{o}_{KN}$ correspond to nuisance factors, which are not used in any puzzle instances.}
  \label{fig:ovector}
\end{wrapfigure}
Ideally, some $K$ factors in $\bm{z}_{ij}$ should be isomorphic to the attributes themselves in this simple RPM setting, after an optimal model is learned.  We index those $K$ relevant factors with a hierarchical indexing model, illustrated in \autoref{fig:ovector}.

The latent attribute selection vector $\bm{o} \in \{0,1\}^K$ determines which, among the $K$ possible, factors are used in the puzzle.  This vector is embedded over a larger attribute-noise selection vector $\bm{o}_{KN} \in \{0,1\}^{K+N}$. In $o_{KN,j} = 0, \quad j=K+1,\ldots,K+N$ indicate the factors corresponding to nuisance, determination of which is a part of the inference process, as defined below in \eqref{eq:indz}.

This latent vector gives rise to ambient images through some stochastic nonlinear mapping $\bm{x} \sim p(\mb{x}|f(\bm{z}|\Theta))$\footnote{We drop RPM indices, where obvious.}, where $\mb{Z} = [ \mb{z}_{ij} ]_{M\times M} \in \Re^{M\times M\times (K+N)}$ is the mapped latent tensor for RPM,  parameterized by $\Theta$ which is to be learned,
\begin{equation}
    \label{eq:obsxz}
    p(\mathcal{M}|\mb{Z},\Theta) = \prod_i\prod_j p(\mb{x}_{ij}|f(\bm{z}_{ij}|\Theta)),
\end{equation}
The RPM inductive bias comes from the way (prior) $\bm{z}_{ij}$ are formed, given the unknown rule $r$.  Specifically,
\begin{equation}
    \label{eq:indz}
    \mb{z}_{ij} = \bm{o}_{KN} \odot
    \begin{bmatrix}
        \bm{o} \odot \mb{z}_{ij,o} + \bm{\overline{o}} \odot \mb{z}_{ij,\overline{o}}  \\
        \mb{0}_{N}
    \end{bmatrix} 
    + 
    \bm{\overline{o}}_{KN} \odot
    \begin{bmatrix}
        \mb{0}_{K} \\
        \mb{z}_{ij,n}
    \end{bmatrix},
\end{equation}
where
$\mb{z}_{ij,o} \in \Re^K$ is the latent representation of the factors that are used in rule $r$, $\mb{z}_{ij,\overline{o}} \in \Re^K$ is the latent representation of the complementary, unused factors\footnote{We use $\mathbf{0}_{l}$ notation for a vector of all zeros of dimension $l$.}.

The key in RPM is to define the priors on factors.  The factors used in the rule, grouped as the tensor $\mb{Z}_o = \left[ \mb{z}_{ij,o} \right]_{M \times M}$ , follow a joint density over  $j=1,\ldots,M$ in row $i$
\begin{equation} \label{eq:ofac}
    p\left( \mb{Z}_o |r \right) = \prod_i p\left( \mb{z}_{i:,o} | r \right) = \prod_i p\left( \mb{z}_{i1,o},\ldots,\mb{z}_{iM,o} | r \right),
\end{equation}
where $\mb{z}_{i:,o}$ is the matrix of size $K \times M$ or all latent representations in row $i$ of RPM.  The factors \emph{not} used in the rule, $\overline{\mb{o}}$, and actors representing the noise information have a different, iid prior (refer inner-most plate,``columns: M'', in \figref{gmrpm})
\begin{small}
\begin{equation}\label{eq:negofac}
    p\left( \mb{Z}_{\overline{o}} |r \right) = \prod_i \prod_j p\left( \mb{z}_{ij,\overline{o}} \right) \text{,} \quad p\left( \mb{Z}_{n} \right) = \prod_m \prod_j \mathcal{N}\left( \mb{z}_{mj,n}; 0, I \right)
\end{equation}
\end{small}
We assume that all $K+N$ factors are independent,
\vspace{-.5em}\begin{small}
\begin{gather*}
    p\left( \mb{z}_{i:,o} | r \right)= \prod_k p\left( \mb{z}^k_{i:,o} | r \right) \text{,} \quad
    p\left( \mb{z}_{ij,\overline{o}} \right) = \prod_k p\left( {z}^k_{ij,\overline{o}} \right) \\ 
    p\left( \mb{z}_{ij,n} \right) = \prod_k p\left( {z}^k_{ij,n} \right).
\end{gather*}
\end{small}

This gives rise to the full Generative Models (\tb{GM-RPM}),
\vspace{-.5em}\begin{multline} \label{eq:gen}
    P(\mathcal{M}; \bm{Z}, \bm{Z}_{o}, \bm{Z}_{\overline{o}}, \bm{Z}_{n}, r, \mb{o}, \mb{o}_{KN}) = \\
    p(\mathcal{M}| \bm{Z}) p( \bm{Z} | \bm{Z}_{o}, \bm{Z}_{\overline{o}}, \bm{Z}_n, \mb{o}, \mb{o}_{KN} ) p( \bm{Z}_{o} | r ) p( \bm{Z}_{\overline{o}} ) p(\bm{Z}_n).
\end{multline}

\paragraph{Inference Model}.
As described in \secref{gmrpm}, the goal is to infer the value of the latent variables that generated the observations, \ie to calculate the posterior distribution over $p(\mb{Z},\mb{o},r|\mathcal{M})$, which is intractable.
Instead, an approximate solution for the intractable posterior was proposed by \cite{kingma2013auto} that uses a variational approximation $q(\mb{Z},\mb{o},r|\mathcal{M};\phi)$, where $\phi$ are the variational parameters.  In this work, we further define this variational posterior as
\vspace{-.5em}\begin{small}
    \begin{multline}
        \label{eq:varpos}
        q(\mb{Z}, \mb{o}, \mb{o}_{KN}, r|\mathcal{M};\phi) \propto 
        q_Z(\mb{Z}| \mb{Z}', r,\mb{o},\mb{o}_{KN} ) q_r(r|\mb{Z}',\mb{o})  \\ q_o(\mb{o} | \mb{o}_{KN}, \mb{Z}') q_o(\mb{o}_{KN} | \mb{Z}') \prod_{ij} q_{\phi}(\mb{z}'_{ij}|\mb{x}_{ij}),
    \end{multline}
\end{small}
where $\mb{Z}'$ is an intermediate variable which is used to arrive at the final estimate of $\mb{Z}$ using the Factor Consistency inference as described further in this section.


Currently, \daren is designed for $r$ = ``constant-in-a-row", i.e., $q_r(r|\mb{Z}',\mb{o}) = \delta(r-r_{const})$ from \eqref{eq:varpos}. Next, we describe the sequential inference process for all the latents.

\paragraph{Infer} $\mb{Z}'$: Intermediate latent factors $\mb{Z}'$ are first inferred independently for each element $\mb{x}_{ij}$ using a general stochastic encoder $q_{\phi}$ of the VAE family:
\begin{equation}
    q_{\phi}(\mb{Z}'|\mathcal{M}) = \prod_{ij} q_{\phi}(\mb{z}_{ij}'|x_{ij}).
\end{equation}
Our framework accepts arbitrary choices of the VAE-family encoders, as discussed in \autoref{sec:daren}.

\paragraph{Infer} $\mb{o}_{KN}$: To infer $K$, we prune out the $N$ nuisance attributes from $\bm{Z}' \in \Re^{K+N}$ that have collapsed to the prior $\big (q_{\phi}(z'_{ij}|x_{ij}) = p(z'_{ij}) \big )$. Thus the remaining latent dimensions form the relevant $K$ attributes. This is similar to computing the empirical variance $\mathbb{V}(\mb{Z}')$ to set the indices of $\mb{o}_{KN}$ where the variance is above a threshold ($\epsilon = 0.05$).
\begin{gather}
    q_o(\mb{o}_{KN} | \mb{Z}') = \delta(\mb{o}_{KN} - \hat{\mb{o}}_{KN}(\mb{Z}')) \\
    \hat{\mb{o}}_{KN}(\mb{Z}') = 
    \mathds{1}^K_{{\mathbb{V}} (\mb{Z}') > \epsilon },
\end{gather}
where $\mathds{1}^K_{a(x) \geq b}$ is the multi-hot indicator vector whose entries are set to $1$ for the $K$ largest values of $a(x)$ for which $a(x) \geq b$ holds.
These $K$ factors model the actual ground truth factors, while the remaining $N$ factors $\mb{z}'_{ij,n}$ are considered nuisances.

\paragraph{Infer} $\mb{o}$:
Next, we use the multi-hot vector $o_{KN}$ to set only the selected attribute for the given instance of RPM.
For a given $\mathcal{M}$ and $r$ = constant in a row, $\mb{o}_{KN,i:}$ values remains the same for images in row $i$. We utilize KL divergence as a measure over all pairwise $z'_{i:}$ and set only on the indices with $l$-lowest divergence values to arrive at
\begin{small}
\begin{gather}
    \label{eq:fixedfactor}
    q_o(\mb{o} | \mb{o}_{KN}, \mb{Z}') = \delta(\mb{o} - \hat{\mb{o}}(\mb{o}_{KN}, \mb{Z}')) \\
    \hat{\mb{o}}(\mb{o}_{KN}, \mb{Z}') = \mathds{1}^l_{ -\boldsymbol{\delta}_{KL} } \\
    \boldsymbol{\delta}_{KL}(k) = \frac{1}{M^3} \sum_{m,i,j=1}^{M} D_{KL} \big ( q_{\phi}(\mb{z}_{ij}^{k}{'} | \mb{x}_{ij}) \lvert \rvert q_{\phi}(\mb{z}_{im}^{k}{'} | \mb{x}_{im}) \big )
\end{gather}
\end{small}
where, for $k=1,\ldots,K$.

\paragraph{Infer $\mb{Z}$ using Factor Consistency:} We describe the process of estimating $\mb{Z}$ from the intermediate variable $\mb{Z}'$ using $q_Z(\mb{Z}|\mb{Z}', r,\mb{o},\mb{o}_{KN} )$, for the chosen case of $r$; the goal here is to obtain consistent, denoised final estimates of the factors, given the intermediate noisy estimates $\mb{Z}'$ and the estimated relevant factors $\mb{o}$.  Specifically,
\begin{small}
\begin{equation} \label{eq:zprimetoz}
        \hat{\mb{z}}_{ij} = \hat{\bm{o}}_{KN} \odot
    \begin{bmatrix}
        \hat{\bm{o}} \odot f_{avg}(\mb{z}'_{i:,o}) + \hat{\bm{\overline{o}}} \odot \mb{z}'_{ij,\overline{o}}  \\
        \mb{0}_{N}
    \end{bmatrix} 
    + 
    \hat{\bm{\overline{o}}}_{KN} \odot
    \begin{bmatrix}
        \mb{0}_{K} \\
        \mb{z}'_{ij,n}
    \end{bmatrix},
\end{equation}
\end{small}
and $q_Z(\mb{Z}|\mb{Z}', r, \mb{o},\mb{o}_{KN} ) = \delta(\hat{\mb{Z}} - \hat{\mb{Z}})$.

Since the relation $r$ acts on the latent vector $\bm{Z}_{o}$, we apply the averaging strategy on it.
For $r=r_{const}$, the averaging strategy is a variant of the method in Multi Level VAE~\cite{bouchacourt2018multi} described for row $i$ as:
\vspace{-.5em}\begin{equation}
    \label{eq:avgfn}
        f_{avg}\big ( \mb{z}'_{ij,o} \big ) =  \frac{1}{M} \sum_{j=1}^{M} \mb{z}'_{ij,o}.
 \end{equation}

Using \eqnref{zprimetoz} \& \eqnref{avgfn}, we obtain updated rule-attribute constrained latent representations $\mb{Z}$ for each in $\mathcal{M}$.
The resulting $\mb{Z}$ is given as an input to the decoder network to learn to reconstruct the original Raven's matrix, which we denote $\hat{\mathcal{M}}$.

\begin{figure}[!ht]
    \begin{center}
        \includegraphics[width=.99\linewidth]{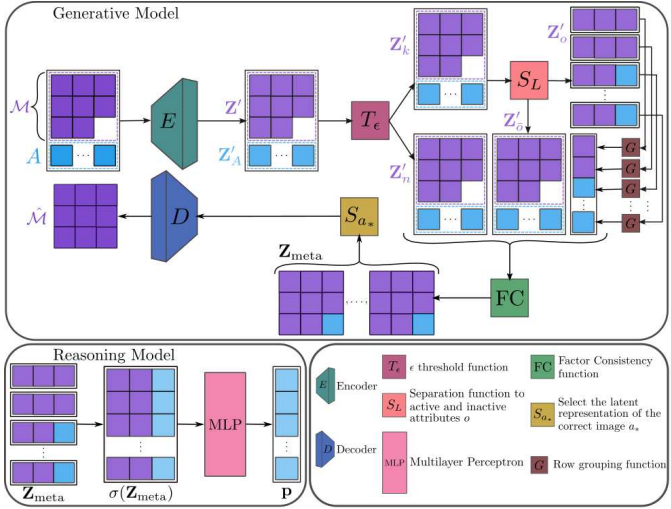}
    \end{center}
    \caption{Illustration of \tb{\daren}. It consists of a VAE-based generative and reasoning network. \tb{Generative Model}. The encoder encodes $\mathcal{M}$ and $A$ to $\mb{Z}'$ and $\mb{Z}'_A$. The  $K$ possible attributes are learned from  $\mathcal{M}$  by picking the factor indices with high variance, and the rest are kept as nuisances ($N$), performed by the threshold function $T_\epsilon$. Next, with $S_L$, we further split the $K$-D latent representation to two groups of vectors $\mb{Z}'_{o}$ and $\mb{Z}'_{\bar{o}}$ related to  the active (set bits of $o$) and inactive (set bits of $\bar{o}$) attributes. No operation is performed on the factor indices at the set bit of $\overline{o}$ and the nuisance factors. The rule constraint is enforced on the set bit of $\bm{o}$ to take the same value via an averaging strategy (G). The decoder recieves the updated latent representation of the correct choice image $a_*$ to reconstruct the image back ($\hat{\mathcal{M}}$). \tb{Reasoning Model}. We consider the latent representation $\mb{Z_\text{meta}}$ to extract the standard deviation across factor index for the top two rows and all possible six rows. An MLP trained on the concatenated standard deviation of the top two rows with choice $a_i$ predicts the best fit image.}
    \label{fig:daren}
\end{figure}
\subsection{\tb{\daren}} \label{sec:daren}
Inspired by \textit{GM-RPM}, we propose a novel framework named \tb{D}isentangling based \tb{A}bstract \tb{Re}asoning \tb{N}etwork (\daren).
Please refer to \figref{daren} for an overview of \daren.
\daren is composed of two primary components, a variational auto encoder (VAE) module and a reasoning module.
Using \secref{gmrpm} described above our variational autoencoder (VAE) learns $q_{\phi}(\mb{Z}|\mathcal{M})$ ($\mb{Z}$ is the final estimate) and $p_{\theta}(\mathcal{M}|\mb{Z})$, where the former is refered as encoder or inference model and the later as decoder or generative model.  We assume that the factors vary independently, hence to drive to maximize statistical independence we append the VAE evidence lower bound objective with the Total Correlation (TC) term \cite{kim2018disentangling},
\begin{small}
\begin{equation}
    TC(\mb{Z}) = \prod_{i}\prod_{j}D_{KL}\big [ q(z_{ij}) \lvert\rvert  \underset{l}{\Pi} q(z_{ij}^{l}) \big ]
\end{equation}
\end{small}
The form of the augmented ELBO objective is described as:
\begin{small}
\begin{equation}
    \begin{split}
        \label{eq:regvae}
        &L_{\phi, \theta}(\mathcal{M})\!=\!
        E_{p(\mathcal{M})} [ E_{{q_{\phi}(\mb{Z}_o|\mathcal{M},r)} {q_{\phi}(\mb{Z}_{\overline{o}}|\mathcal{M},r)} {q_{\phi}(\mb{Z}_n|\mathcal{M})}} 
        [ \log p_{\theta}(\mathcal{M}|\mb{Z}) ] ] \\
        & - \lambda_1 \big [ D_{KL}[ q(\mb{Z}_o|\mathcal{M},r) \lvert\rvert p(\mb{Z}_o|r)  ] + D_{KL}[ q(\mb{Z}_{\overline{o}}|\mathcal{M},r) \lvert\rvert  p(\mb{Z}_{\overline{o}}|r)  ] \\
        & + D_{KL}\big [ q(\mb{Z}_n|\mathcal{M}) \lvert\rvert p(\mb{Z}_n)  ] \big ]
        - \lambda_2  TC(\mb{Z}),
    \end{split}
\end{equation}
\end{small}
where hyperparameter $\lambda_1, \lambda_2$ controls the weight on the KL-divergence and the TC  respectively.

\paragraph{Reasoning Module}.\label{sec:reasoning}
The reasoning component of \daren incorporates a relational structure to infer the abstract relationship on the attribute $\mb{o}$ for images in $\mathcal{M}$.
The reasoning module receives disentangled representations, $\mb{Z}$ (of $\mathcal{M}$) and $\mb{Z}_A$ (of choices $A$).
We prepare $\mb{Z_{\text{meta}}}$ of size $(M + |A| -1) \times M$ by iteratively filling each choice as the missing piece. 


Therefore, solving $\mathcal{M}$ is equivalent to finding the correct row in $\{M, \cdots, M+|A|-1\}$, that satisfies the same rules shared by the top $M-1$ rows.
We compute the variance for all $M$ representation in each row $i$, $\sigma(\mb{Z_{\text{meta}}}_{(i)})$ over each dimension in the latent representation.
The above process is applied to all the  $M + A - 1$ rows that include all $|A|$ probable last rows.
Next, we concatenate the variance vector of the top $M-1$ rows with each probable last row (\figref{daren}) to prepare $|A|$ choice variance vectors, $\mathbb{R}^{|A| \times M \times (K+N)}$.
We feed this concatenated variance vector to a three-layered MLP ($\psi$) with a ReLU and a dropout layer \cite{srivastava2014dropout}, the probability of $\overline{a}$ is estimated as:
\begin{small}
\begin{equation}
    p_i = \psi([\sigma(\mb{Z_{\text{meta}}}_{(1)})\  \cdots\  \sigma(\mb{Z_{\text{meta}}}_{(M-1)})\  \sigma(\mb{Z_{\text{meta}}}_{(i)})]),
\end{equation}
\end{small}
where $i \in \{M, \ldots, (M+|A|-1)\}$ corresponds to $|A|$ probable choices. The choice with the highest score is predicted as the correct answer. The above process is trained using a Cross Entropy loss function.

\begin{table*}
\vspace{0.3cm}
\begin{minipage}{0.58\linewidth}
\centering
\begin{adjustbox}{width=\linewidth}
\begin{tabular}{|c|c|c|cccc|}
        \toprule
        \tb{Dataset} & \tb{Model}           & \tb{Reasoning}                &                               & \tb{Disentanglement}          &                               &                              \\
                     &                      &                               & \tb{F-VAE}                    & \tb{DCI}                      & \tb{MIG}                      & \tb{SAP}                     \\
        \midrule
        \addlinespace[0.1cm]
        \multirow{3}{*}{\tb{DSprites}}
                     & \textit{Staged-WReN} & 97.4 $\pm$ 4.2                & 74.4 $\pm$ 7.3                & 52.9 $\pm$ 10.5               & 28.7 $\pm$ 11.5               & 4.0 $\pm$ 1.4                \\
                     & E2E-WREN             & \red{99.6} $\pm$ \red{0.5}    & \blue{77.6} $\pm$ \blue{5.0}  & \blue{58.1} $\pm$ \blue{8.0}  & \blue{38.2} $\pm$ \blue{7.7}  & \red{6.0} $\pm$ \red{2.2}    \\
                     & \tb{\daren}          & \red{99.3} $\pm$ \red{0.5}    & \red{79.2} $\pm$ \red{6.2}    & \red{59.0} $\pm$ \red{6.4}    & \red{39.0} $\pm$ \red{0.0}    & \red{6.0} $\pm$ \red{2.0}    \\
        \addlinespace[0.1cm]
        \midrule
        \addlinespace[0.1cm]
        \multirow{3}{*}{\tb{Mod Dsprites}}
                     & \textit{Staged-WReN} & 80.0*                         & 44.0 $\pm$ 10.6               & 31.2 $\pm$ 7.6                & 13.8 $\pm$ 7.1                & 6.4 $\pm$ 2.6                \\
                     & E2E-WREN             & \blue{85.7} $\pm$ \blue{11.4} & \blue{65.1} $\pm$ \blue{10.6} & \blue{43.0}  $\pm$ \blue{6.9} & \blue{26.1} $\pm$ \blue{9.1}  & \blue{8.3} $\pm$ \blue{3.3}  \\
                     & \tb{\daren}          & \red{86.5} $\pm$ \red{10.4}   & \red{77.0} $\pm$ \red{13.2}   & \red{50.1} $\pm$ \red{11.3}   & \red{34.9} $\pm$ \red{12.4}   & \red{12.6} $\pm$ \red{4.8}   \\
        \addlinespace[0.1cm]
        \midrule
        \addlinespace[0.1cm]
        \multirow{3}{*}{\tb{Shapes3D}}
                     & \textit{Staged-WReN} & 90.0*                         & 84.5 $\pm$ 8.7                & 73.9 $\pm$ 9.0                & 44.6 $\pm$ 8.8                & 6.3 $\pm$ 2.9                \\
                     & E2E-WREN             & \blue{98.3} $\pm$ \blue{2.0}  & \blue{91.3} $\pm$ \blue{6.5}  & \blue{79.1}  $\pm$ \blue{7.7} & \blue{54.9} $\pm$ \blue{15.7} & \blue{8.4} $\pm$ \blue{3.8}  \\
                     & \tb{\daren}          & \red{99.2} $\pm$ \red{0.8}    & \red{98.4} $\pm$ \red{3.2}    & \red{91.6} $\pm$ \red{4.7}    & \red{68.8} $\pm$ \red{17.5}   & \red{17.2} $\pm$ \red{4.8}   \\
        \addlinespace[0.1cm]
        \midrule
        \addlinespace[0.1cm]
        \multirow{3}{*}{\tb{Realistic}}
                     & \textit{Staged-WReN} & 55.5 $\pm$ 8.1                & 45.0 $\pm$ 5.5                & 37.4 $\pm$ 4.6                & 22.7 $\pm$ 7.7                & 9.8 $\pm$ 2.6                \\
                     & E2E-WREN             & \blue{72.7} $\pm$ \blue{7.5}  & \blue{55.3} $\pm$ \blue{6.1}  & \blue{42.5}  $\pm$ \blue{6.4} & \blue{29.7} $\pm$ \blue{8.1}  & \blue{12.8} $\pm$ \blue{2.9} \\
                     & \tb{\daren}          & \red{73.5} $\pm$ \red{6.5}    & \red{75.8} $\pm$ \red{8.6}    & \red{50.6} $\pm$ \red{5.9}    & \red{36.1} $\pm$ \red{8.2}    & \red{19.3} $\pm$ \red{5.2}   \\
        \addlinespace[0.1cm]
        \midrule
        \addlinespace[0.1cm]
        \multirow{3}{*}{\tb{Real}}
                     & \textit{Staged-WReN} & 61.6 $\pm$ 9.4                & 57.8 $\pm$ 7.5                & 46.1 $\pm$ 2.5                & 31.2 $\pm$ 6.1                & 14.4 $\pm$ 4.0               \\
                     & E2E-WREN             & \blue{72.2} $\pm$ \blue{8.8}  & \blue{65.0} $\pm$ \blue{7.3}  & \blue{48.0}  $\pm$ \blue{3.5} & \blue{34.3} $\pm$ \blue{7.0}  & \blue{18.2} $\pm$ \blue{4.4} \\
                     & \tb{\daren}          & \red{74.3} $\pm$   \red{9.9}  & \red{75.8} $\pm$ \red{9.7}    & \red{51.8} $\pm$ \red{4.9}    & \red{37.0} $\pm$ \red{8.1}    & \red{20.8} $\pm$ \red{5.3}   \\
        \addlinespace[0.1cm]
        \midrule
        \addlinespace[0.1cm]
        \multirow{3}{*}{\tb{Toy}}
                     & \textit{Staged-WReN} & 64.2 $\pm$ 13.3               & 49.2 $\pm$ 4.1                & 43.0 $\pm$ 2.5                & 29.7 $\pm$ 7.0                & 10.6 $\pm$ 2.4               \\
                     & E2E-WREN             & \blue{80.8} $\pm$ \blue{3.5}  & \blue{58.6} $\pm$ \blue{4.2}  & \blue{48.0}  $\pm$ \blue{2.9} & \red{37.8} $\pm$ \red{7.9}    & \blue{13.7} $\pm$ \blue{2.9} \\
                     & \tb{\daren}          & \red{81.5} $\pm$ \red{6.8}    & \red{75.3} $\pm$ \red{15.4}   & \red{52.8} $\pm$ \red{6.8}    & \blue{35.1} $\pm$ \blue{9.6}  & \red{18.7} $\pm$ \red{5.5}   \\
        \addlinespace[0.1cm]
        \bottomrule
    \end{tabular}
    \end{adjustbox}
\caption{\hspace*{0mm}Performance (mean $\pm$ variance) of \tb{Reasoning} accuracy and four widely used benchmark \tb{Disentanglement Metrics} on the six benchmark datasets. Note: higher score implies better result. The best score for each dataset among the competing models are shown in \red{bold red} and second-best in \blue{blue}. (Note: $^*$ values are taken from~\protect\cite{van2019disentangled}.)}\label{tab:reas_dis_results}
\end{minipage} \hfill
\begin{minipage}{0.39\linewidth}
\centering
\includegraphics[width=1.0\linewidth]{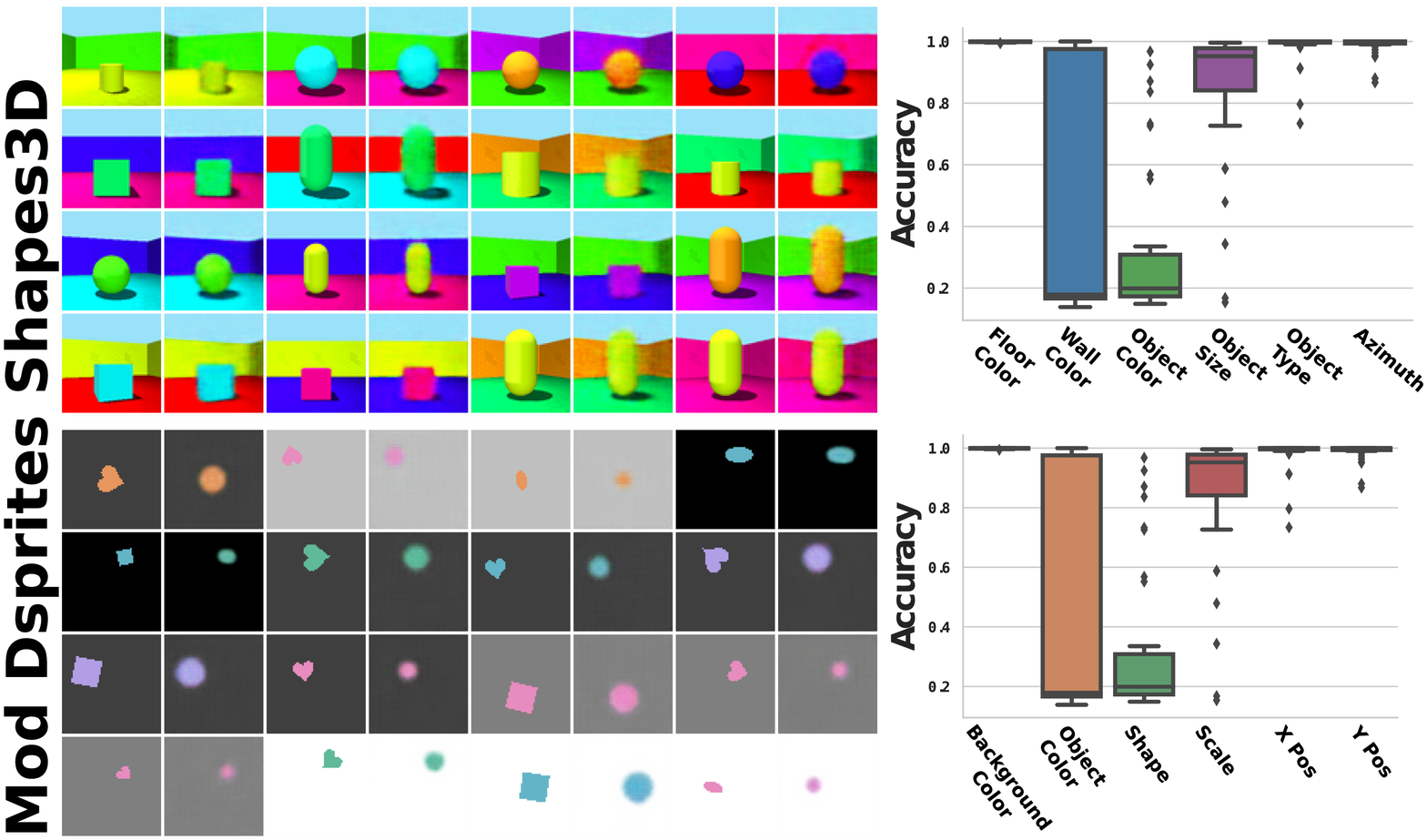}
\captionsetup{type=figure} 
\caption{Top: \tb{Shapes3D}, Bottom: \tb{Mod DSprites}. In the order of left to right, Reconstructions from \daren  (representative samples of median reconstruction error), odd columns show real samples and even columns their reconstruct. Distribution of reasoning performance per generative attribute over all 35 trained models. Expected prior KL-divergence for individual dimensions (left plot represents the best performing model, right plot represents the lowest performing model).}\label{fig:qual}
\end{minipage}
\end{table*}

\section{\tb{Experiments}}
\label{sec:exp}

\subsection{\tb{Datasets, Baselines, Experimental Setup}} \label{rpm_dataset}

We study the performance of \daren on six benchmark datasets used in disentangling work,  (i) \textit{dsprites} \cite{matthey2017dsprites}, (ii) modified \textit{dSprites}~\cite{van2019disentangled}, (iii) \textit{shapes3d}~\cite{kim2018disentangling}, (iv-vi) MPI3D -- (Real, Realistic, Toy)~\cite{gondal2019transfer} .
We use experimental settings similar to~\cite{van2019disentangled} to create RPM for the above datasets.
The training procedure in~\cite{van2019disentangled}, referred as \textit{Staged-WReN}, is used as our baseline model.
The \textit{Staged-WReN} is a two-stage training process, where a disentangling generative network is first trained ( $\sim300K$ iterations), followed by training a Wild Relational Network (WReN)~\cite{barrett2018measuring} on RPM using the representation (at $300K$) obtained from the trained encoder.
Building on \textit{Staged-WReN}, we propose an adapted baseline referred to as \textit{E2E-WReN}, where we jointly train both the disentangling and the reasoning network from end-to-end.
To train \daren, we use a warm start by initializing only its VAE parameters with a partially trained Factor VAE~\cite{kim2018disentangling} model for $\sim100K$ on the above datasets (not RPM instances) followed by training \daren for $200K$ iterations on the RPM generated from the datasets. We evaluate the model's performance on fresh RPM samples\footnote{The state space of all RPM questions is huge e.g. in \textit{shapes3d} has $10^6$ factor combinations possible per image and a total of 14 total images for M=3, A=6 yielding $\sim 10^{6^{14}}$ possible RPM (minus invalid configurations)}. Refer to Appendix for details on the dataset, experimental setup, \textit{E2E-WReN}, and additional results.

\subsection{\tb{Evaluating Abstract Visual Reasoning Results}}
\label{sec:exp_reas}
We compare our results in both the \tb{reasoning accuracy} and the \tb{disentanglement scores} against the SOTA methods.  Both scores are important in the context of interpretable reasoning models, as the reasoning accuracy alone does not necessarily reflect the discovery of the underlying RPM factors.  Both scores are shown in \autoref{tab:reas_dis_results}.

\paragraph{\tb{Reasoning}.} We report the performance  of reasoning accuracy in the column ``Reasoning'' in \tabref{reas_dis_results}.
Our proposed model \daren, compared against \textit{Staged-WReN} and \textit{E2E-WReN}, shows an improvement of $\bm{\sim 2-17} \% $  and $\bm{\sim 0-2} \%$ respectively.
It is seen from \tabref{reas_dis_results}, our model outperforms the prior work, i.e. \textit{Staged-WReN} by a margin of $\bm{\sim 2-17} \%$, especially for datasets with color as a ground truth factor (all excluding \textit{dsprites}).
The VAE model \daren can firmly separate color attributes from other factors compared to \textit{Staged-WReN}.


The WReN module is similar for both \textit{E2E-WReN} vs. \textit{Staged-WReN}, where it learns a relational structure by aggregating all pairwise relations on the latent space within the images in $\mathcal{M}$ and between $\mathcal{M}$ and candidates in $A$. However, a joint optimization of reasoning + representation (\textit{E2E-WReN}) learns to solve reasoning task better than \textit{Staged-WReN}. Despite the improvement via joint optimization, WReN performance is still sub-optimal in learning the underlying reasoning patterns. For each choice filled RPM, the pairwise relation tensor ($6 \times 9 \times 9$) contains intra-pairwise relations formed within $\mathcal{M}$ ($: \times 8 \times 8$) and rest are inter-pairwise relations between $\mathcal{M}$ and candidates in $A$.
The intra-pairwise relations remain invariant across all six choices. Only the inter-pairwise score between $\mathcal{M}$ and candidates in $A$ play a key role in inferring the relationship. We verify by extracting the output of edge MLP, i.e., $9 \times 9$ feature representations ($\Re^d$), where $d$ is 256 or 512, and computing the L2-norm on these feature vectors. The features of inter pairwise relations given to MLP determine the correct answer. The above is verified on both the trained networks (\textit{Staged-WReN} and \textit{E2E-WReN}) over all the datasets. 
\daren avoids forming redundant relations; instead, it works by matching the attributes to find the correct row that satisfies the rule in the top two rows.
Our results on these datasets provide evidence of a stronger affinity between reasoning and disentanglement (discussed in the section below), which results from jointly learning both tasks.

In \figref{qual}, we present qualitative analysis of \textit{shapes3d} and modified \textit{dsprites}. The quality of the reconstructed images confirms that the learned distribution correctly captured the image content. In the right column, we present the distribution of per attribute accuracy over the hyperparameter sweep. The performance of object color in modified \textit{dsprites} is sensitive to hyperparameter changes which is evident from large variance. We also see low performance for object shape due to its discrete nature; since our current latent representation is modeled towards real values, it fails to handle discrete representation. 

\paragraph{\tb{Disentangling}}. In \tabref{reas_dis_results}, ``Disentanglement'', we report the disentanglement scores of trained models on four widely used evaluation metrics namely,  the Factor-VAE metric~\cite{kim2018disentangling}, DCI~\cite{eastwood2018framework}, MIG~\cite{chenisolating}, and SAP-Score~\cite{kumar2017variational}.
\daren improves the reasoning performance and also strongly disentangles the latent vector, in contrast to both \textit{Staged-WReN} and \textit{E2E-WReN}.
One primary reason for significant improvements by \daren compared to \textit{Staged-WReN} and even \textit{E2E-WReN} is due to the extraction of the $K$ underlying generative factors and the averaging strategy over the least varying index in $K$.
Generally, in an unsupervised training process, weak signals from the true generative factors often leak into nuisance factors. However, \daren avoids such infusion of nuisance factors by separating $o$ from $n$.

\section{\tb{Conclusion}}
\label{sec:conclusion}
We proposed the \textit{GM-RPM} prior and an accompanying learning framework, \daren, to exploit the weak inductive bias present in RPM for visual reasoning. \textit{GM-RPM} provides a backdoor guide on the constraints that can be exploited to solve RPM-based reasoning tasks. To this end, we lifted the emerging evidence of dependency between disentanglement and reasoning~\cite{van2019disentangled} one step further by showing that joint end-to-end learning with the appropriate inductive bias can lead to effective simultaneous representation and reasoning models. 
\daren achieves state-of-the-art results on both the reasoning and the disentanglement metrics. In addition, \daren offers the added flexibility of using arbitrary state-of-the-art factor inference approaches based on ELBO-like objectives to infer and learn the representation needed for accurate reasoning. Our results show evidence of a strong correlation between learning disentangled representation and solving the reasoning tasks. As such, \textit{GR-RPM} and \daren offer a general framework for research in joint learning of representation and reasoning.

\else

\appendix
\subsection{\tb{Experimental Details}}

We use the same experimental setup such as the architecture, hyperparameters used in the prior work~\cite{van2019disentangled}. 
Our VAE network architecture is similar to Factor VAE, and the reasoning model is a three-layered MLP ($\psi(\cdot)$) with a ReLU, and a dropout layer \cite{srivastava2014dropout}.
The architecture details for \daren are depicted in \tabref{arch}.
All our models are implemented in PyTorch~\cite{NEURIPS2019_9015} and optimized using ADAM optimizer~\cite{kingma2014adam}, with the following parameters: learning rate of $1e-4$ for the reasoning + representation network excluding the Discriminator (approximation of TC term in Factor VAE) which is set to $1e-5$, $\beta_1 = 0.9$, $\beta_2=0.999$, $\epsilon=10^{-8}$. To demonstrate that our approach is less sensitive to the choice of the hyper-parameters ($\gamma$), and network initialization, we sweep over 35 different hyper parameter settings  of $\gamma \in \{ 1.0, 10.0,20.0,30.0,40.0,50.0,100.0\}$ and initialization seeds $\{1,2,3,4,5\}$.
The image size used in all six datasets is $64\times 64\times ch$ pixels, where channels, ch$ =3$ for all datasets except for \textit{dsprites} where it is one.
We scaled the pixel intensity to $[0,1]$.
For each model, we use a batch size of $64$ during the training process, where each mini-batch consists of generated random instances RPMs.

$\textit{Staged-WReN}$ used SOTA generative models for disentanglement trained upto $300K$ to train the relational network for $100K$ (VAE parameters frozen).
The relational network is similar to WReN, composed of two MLPs: edge MLP that learns on all pairwise joined representation for each choice, graph MLP that maps the pairwise all feature vector to a single probability score for each choice.
On the other hand for \textit{E2E-WReN}, we trained both the representation (only on Factor VAE) and reasoning model upto $300K$. 
Note: VAE was trained on all factor data and the reasoning network on RPM created by quantized factors similar to ~\cite{van2019disentangled}\footnote{The quantization ensured visually distinguishable factor values for each factor of variation to humans.}.

\begin{table}[ht]
    \centering
    \caption{\hspace*{0mm} Representation (Encoder, Decoder) and Reasoning Architectures.
    }
    \label{tab:arch}
    \small
    \begin{tabularx}{\linewidth}{ll}
        \toprule
        \tb{Encoder}                               & \tb{Decoder}                              \\
        \midrule
        Input: \# channels $\times$ 64 $\times$ 64 & Input  $\Re^{10}$                  \\

        4 $\times$ 4 conv. 32 ReLU. stride 2       & FC. 256 ReLU                              \\
        4 $\times$ 4 conv. 32 ReLU. stride 2       & FC. 4 $\times$ 4 $\times$ 64 ReLU         \\
        4 $\times$ 4 conv. 64 ReLU. stride 2       & 4 $\times$ 4 upconv. 64 ReLU. stride 2    \\
        4 $\times$ 4 conv. 64 ReLU. stride 2       & 4 $\times$ 4 upconv. 32 ReLU. stride 2    \\
        FC 256. FC 2 $\times$ 10                   & 4 $\times$ 4 upconv. 32 ReLU. stride 2    \\
                                                   & 4 $\times$ 4 upconv. \#channels. stride 2 \\
        \bottomrule
    \end{tabularx}
\end{table}

\begin{table}[ht]
    \centering
    \begin{tabularx}{\linewidth}{lc}
        \toprule
        Discriminator       & Reasoning Network                                                                \\
        \midrule
        FC. 1000 leaky ReLU & Input $\mathcal{M}$: $\Re^{8 \times 10}$. $A$: $\Re^{6 \times 10}$ \\
        FC. 1000 leaky ReLU & RN Emb. Size: 54                                                                 \\
        FC. 1000 leaky ReLU & RN MLP: [ 512, ReLU,                                                             \\
        FC. 1000 leaky ReLU & 512, ReLU,                                                                       \\
        FC. 1000 leaky ReLU & 512, ReLU                                                                        \\
        FC. 1000 leaky ReLU & dropout: 0.5,                                                                    \\
        FC. 2               & 1 ]                                                                              \\
        \bottomrule
    \end{tabularx}
\end{table}

\subsection{\tb{Dataset Details}} \label{sec:dataset}

\paragraph{DSprites}. The most commonly used dataset for bench marking disentangling is the \textit{dsprites} dataset~\cite{matthey2017dsprites}.
It consists of a single dsprite (color: white) on a blank background and can be fully described
by five generative factors: shape (3 values), position x (32 values), position y (32 values), size (6 values) and orientation (40 values).
The total images produced are the Cartesian product of the generative factors \ie 737,280.
Ground truth factor changes in the construction of RPMs: three equally distant values of size (instead of 6), four values of x/y position (instead of 32). We ignore orientation factor as certain objects such as squares and ellipses exhibit rotational symmetries. And no changes to shape.

\paragraph{Modified DSprites}. ~\cite{van2019disentangled} is an extension of \textit{dsprites} dataset that adds a background color (5 values), a color attribute (6 values) to the single dsprite, and excluding the orientation factor.
Colors are sampled from a linear HUSL hue space.
The total images produced are the Cartesian product of the generative factors, \ie 552,960.
Ground truth factor changes in the construction of RPMs: shape, size, x/y position remains the same as in \textit{dsprites} with two additional color factors.

\paragraph{3D Shapes}. ``3D Shapes"~\cite{kim2018disentangling} consists of images of a single 3D object in a room and is fully specified by six generative
factors: floor colour (10 values), wall colour (10 values), object colour (10 values), size (8 values), shape (4 values) and rotation (16 values).
The total images produced are the Cartesian product of the generative factors, \ie 480,000.
Ground truth factor changes in the construction of RPMs: four equally distant factors for rotation and size (instead of 8 and 16) and no changes to the rest.

\paragraph{MPI3D--(Real, Realistic, Toy)}. \cite{gondal2019transfer} created a recent bench marking robotics dataset for disentanglement across simulated to real-world environments (Toy, Realistic, Real).
It consists of images of a sprite on a robotic arm and is fully specified by six generative seven generative factors: object color (	6 values), object shape (6 values), object size	(2 values), camera height (3 values), background color (3 values), horizontal axis (40 values), vertical axis (40 values).
The horizontal and vertical axes are sampled from a uniform distribution, and the other five factors take discrete values.
Each dataset contains all possible combinations of the factors that amount to total images in the dataset \ie 1,036,800.
Ground truth factor changes in the construction of RPMs: four equally distant factors for horizontal and vertical axis (instead of 40) and no change to the rest.

\paragraph{Generating RPM Questions.}
Here we describe the process followed to automatically generate RPM, derived from ~\cite{van2019disentangled,wang2015automatic}.
The procedure requires access to the ground truth factors of the dataset.
The dataset generation process can be divided into three parts: i) determining the rule for a sample instance, ii) placing images satisfying the rule in the matrix, iii) generating incorrect alternatives to form choice panel.
We fixed the matrix size as a 3 $\times$ 3 where the bottom right image is removed and placed among five incorrect answers in a randomly chosen location.
We also fixed the relation type to constant in a row.
This means that the three images present in any matrix row should have at least one object factor constant.
In step one of the process, we start by sampling a subset of ground truth factors that we fix across all three rows.
Next, we uniformly sample values without replacement for this fixed subset of factors for each row separately.
The subset of factors is our underlying rule for this sample, where the values taken by them are constant in a row, and this subset of factors remains fixed across all the rows.
Finally, the rest of the factors that are not part of the rule are allowed to take any factor value as long as these values are not the same across the first two rows.
This is done to ensure that each of these rows have a different set of constant factors that are not part of the rule.
However, the solution to the final row must come from the subset of factors that are constant across the first two rows.
Once all the factors are sampled for the $3 \times 3$ matrix, the next step is to retrieve corresponding images from the dataset.
The final step consists of sampling hard incorrect answers for the sample to form the choice panel.
We start by taking the correct answer factors and keep on resampling until it no longer satisfies the underlying rule.
This process is repeated five times to generate alternative panels.
Finally, the correct answer panel is inserted in this choice list in a random position.
Note: The sampled factors are from the list of quantized factors described in \secref{dataset}.
Please refer to \figref{rpm_sample} for sample instances of abstract visual reasoning tasks on all six datasets.

\begin{figure*}[!ht]
\begin{subfigure}{\linewidth}
\centering
\includegraphics[width=0.49\linewidth]{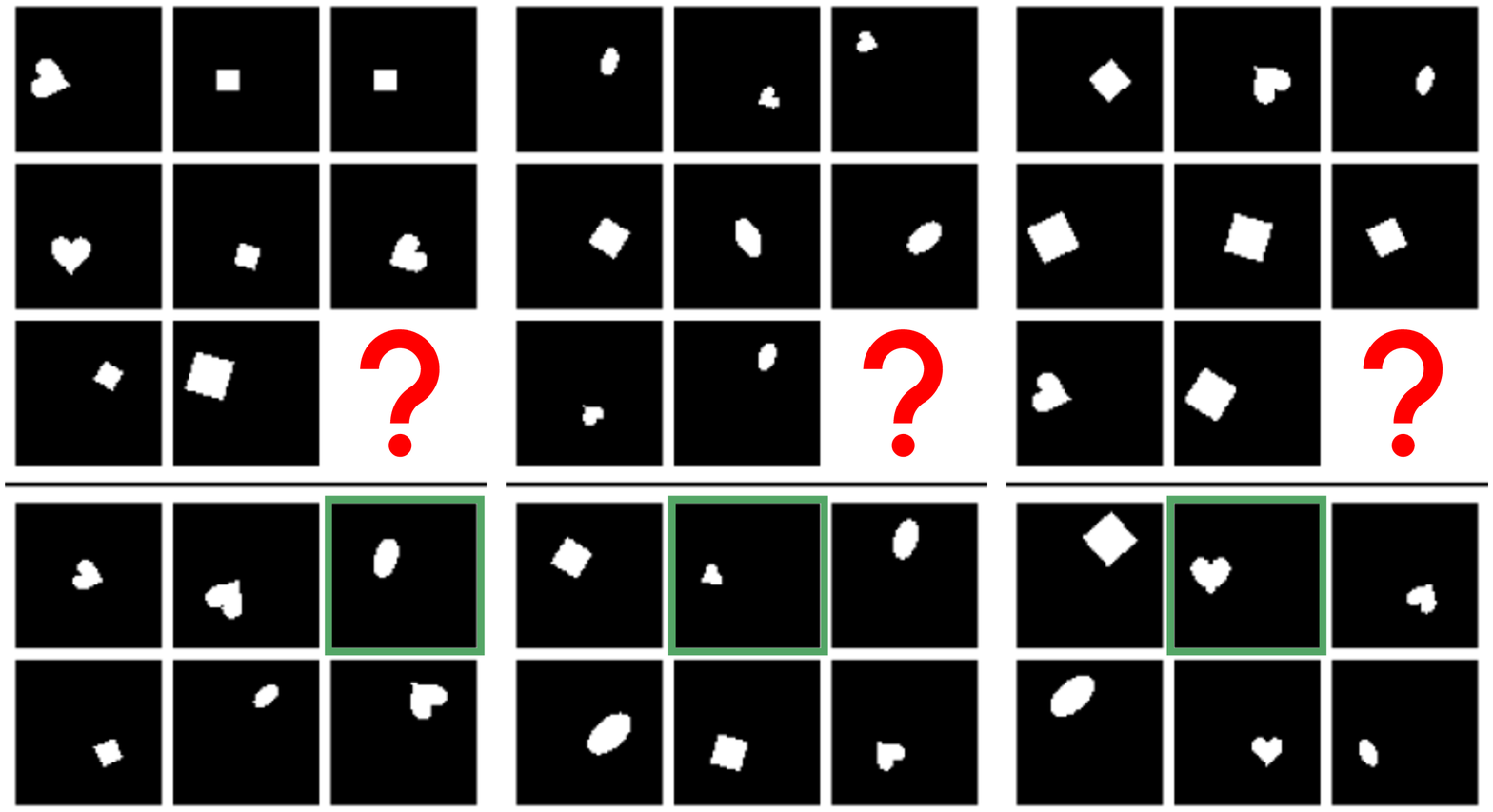}
\hfill
\includegraphics[width=0.49\linewidth]{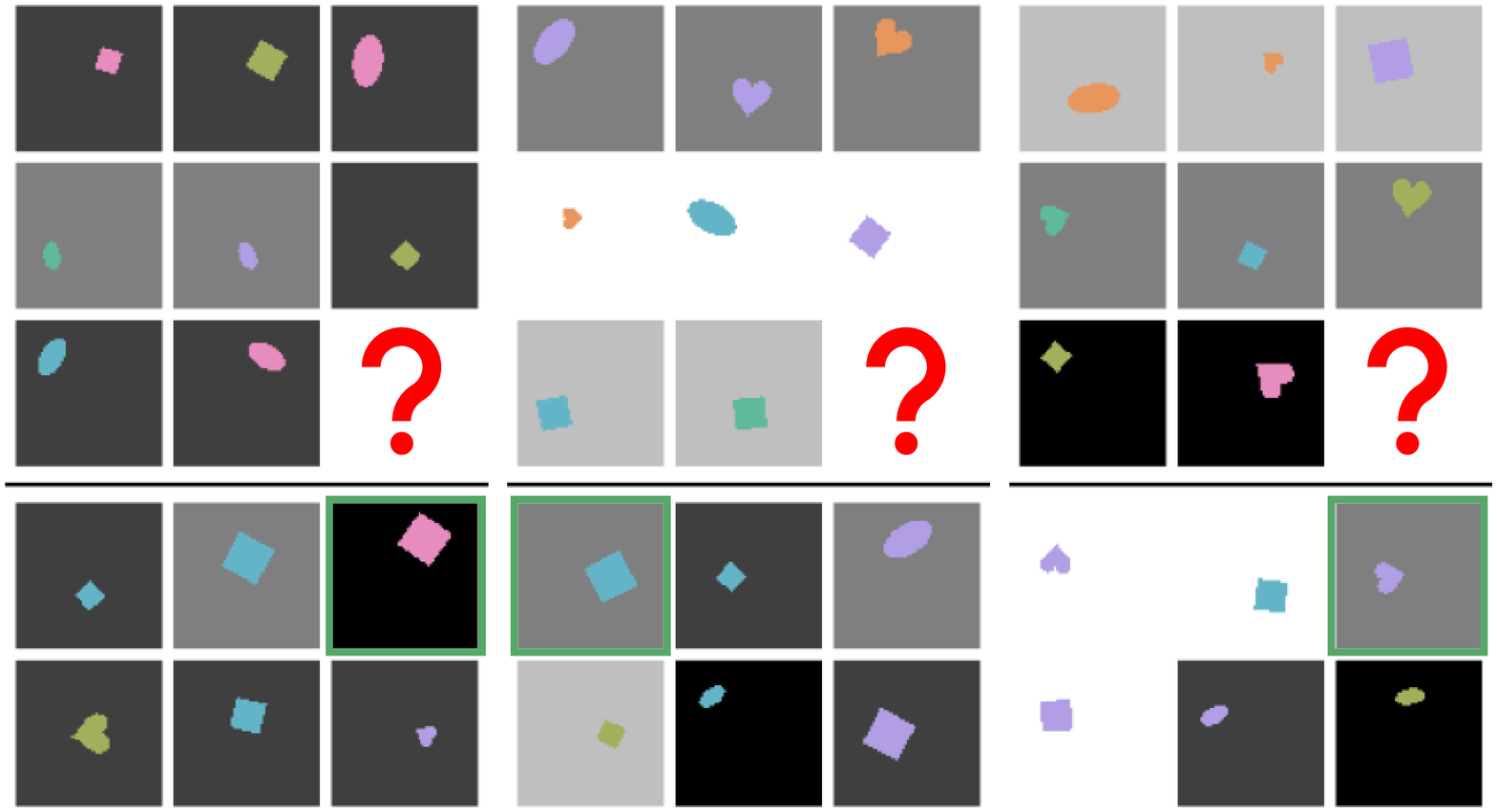}
\end{subfigure} 
\vskip\baselineskip
\begin{subfigure}{\linewidth}
\centering
\includegraphics[width=0.49\linewidth]{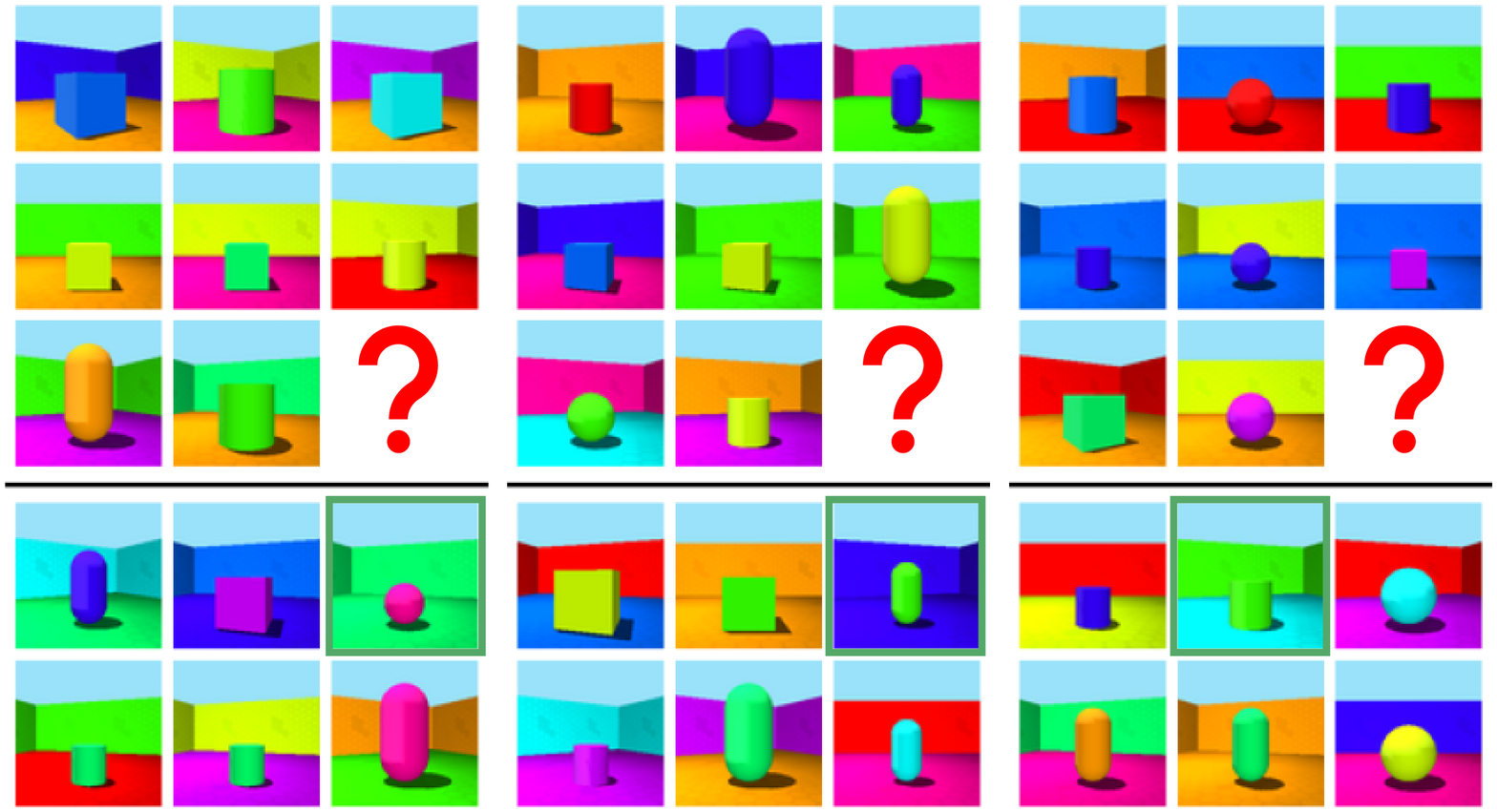}
\hfill
\includegraphics[width=0.49\linewidth]{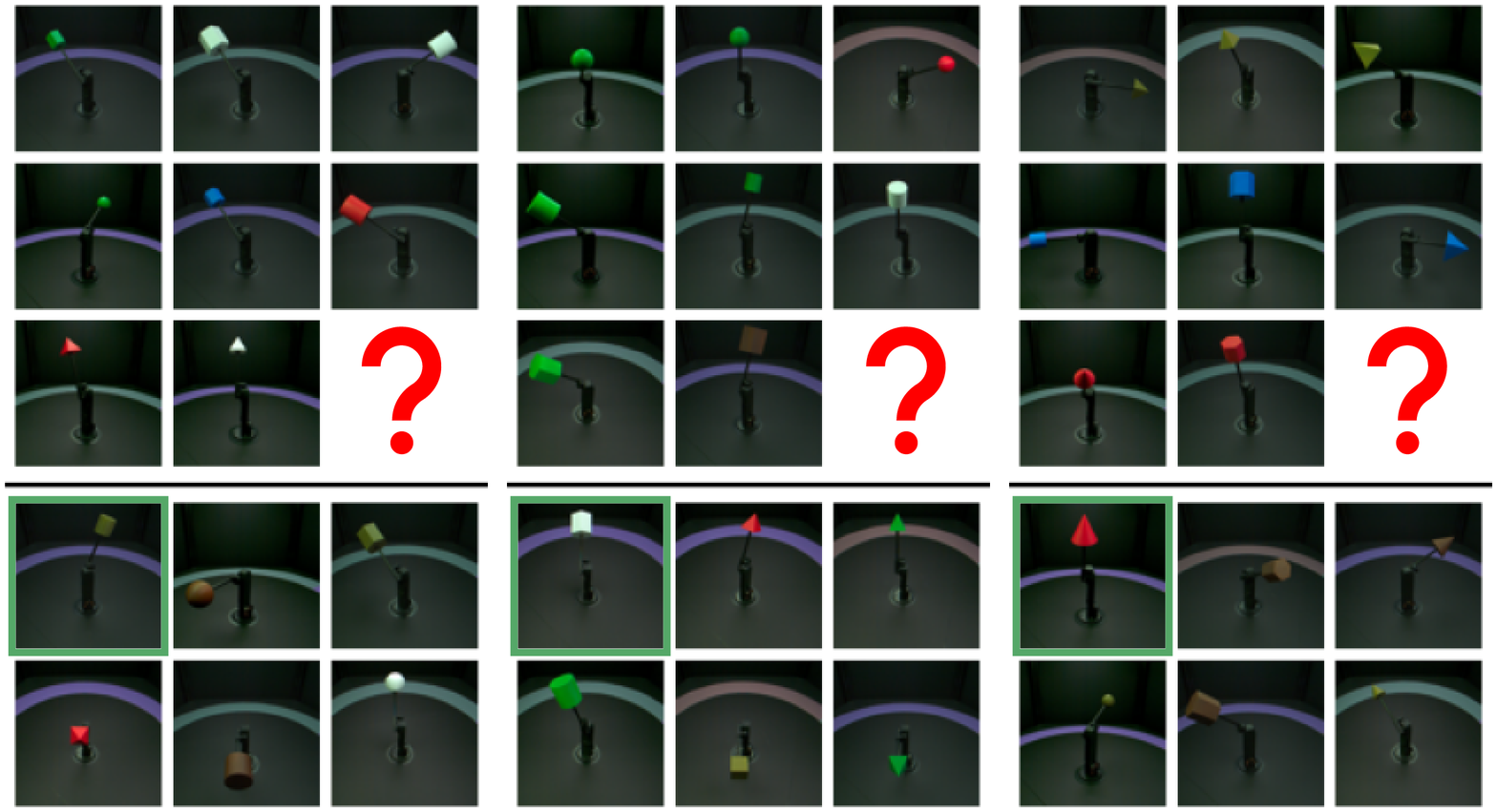}
\end{subfigure} 
\vskip\baselineskip
\begin{subfigure}{\linewidth}
\centering
\includegraphics[width=0.49\linewidth]{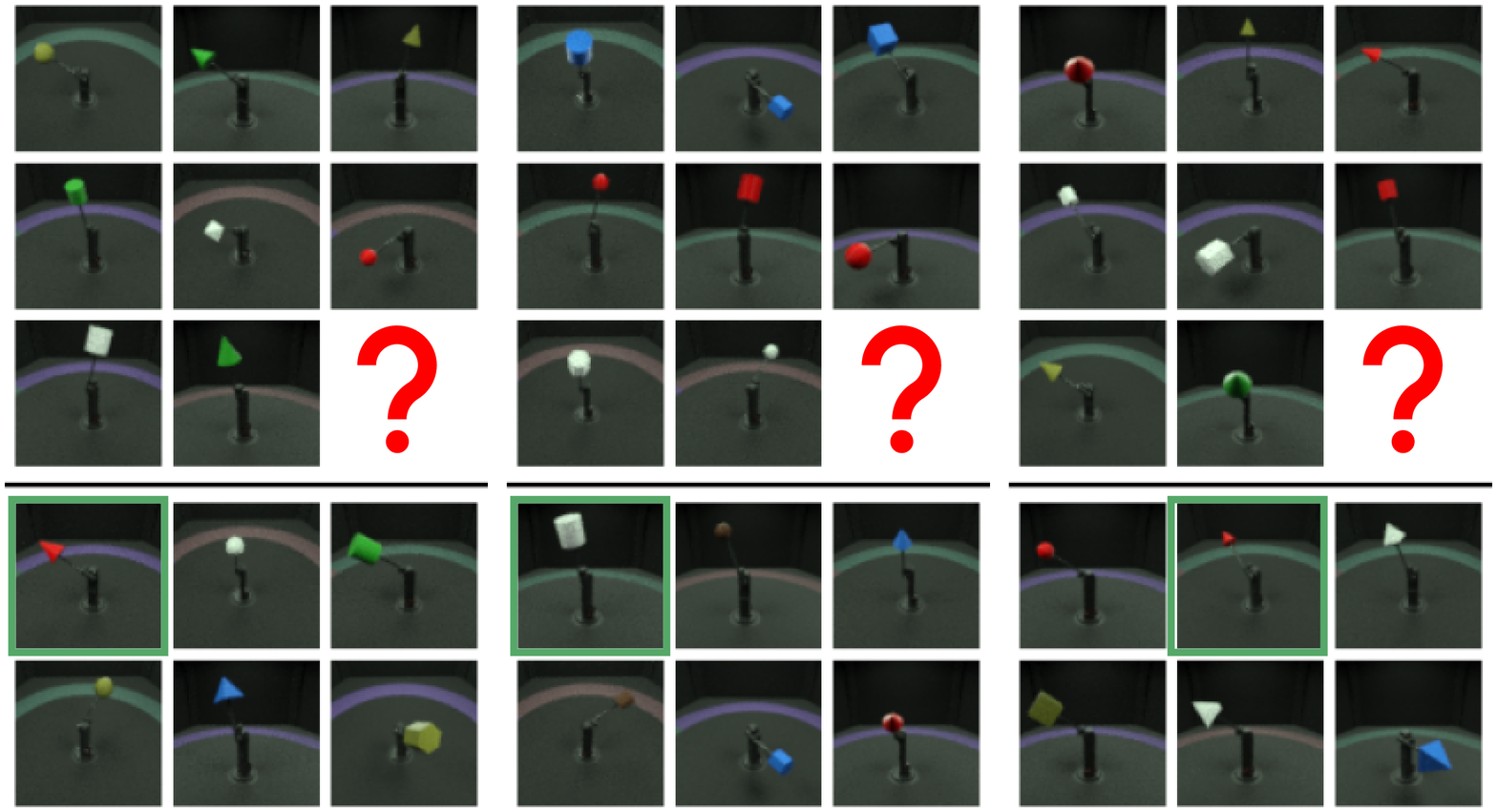}
\hfill
\includegraphics[width=0.49\linewidth]{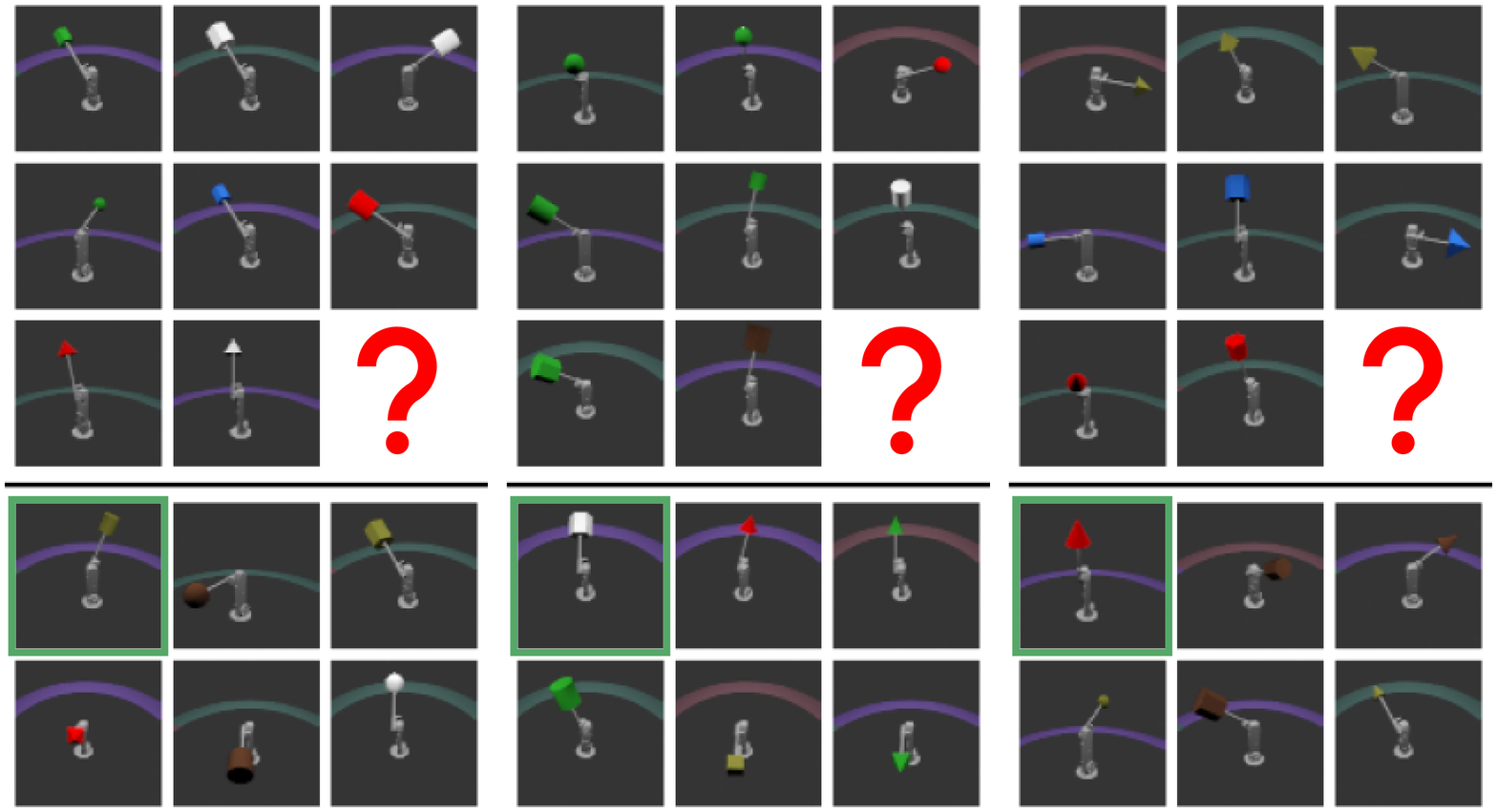}
\end{subfigure} 
\caption{Sample RPM based Abstract Visual Reasoning Tasks for DSprites, Modified DSprites, Shapes3D, MPI3D Real, MPI3D Realistic, MPI3D Toy. (Order: Left to Right, Top to Bottom)}\label{fig:rpm_sample}
\end{figure*}

\subsection{\tb{E2E-WReN}}
For a RPM of size $3 \times 3$, we adopt the same notation used in~\cite{van2019disentangled} to represent the matrix without the bottom last element as $\mathcal{M}$ and the image choices with the correct answer as  the answer panels $ A$ (size = 6 as in prior work).

WReN is evaluated on the embeddings from a deep Convolutional Neural Network (CNN).
For each choice image placed at the missing location, WReN forms all pairwise combination among all the nine latent embeddings and outputs a score.
The relational network is a non-linear composition of two functions $f_{\phi}$ and $g_{\theta}$ which are implemented as multilayer perceptrons (MLPs).
The pairwise embeddings generated above are given as an input to $g_{\theta}$ of the relational network.
This stage is tasked with inferring relations between object properties for a given pair, between context and choice panel, and between context panels.
The output of $g_\theta$ for each pairwise embeddings is aggregated.
The aggregated representation for all six choices is given as an input to the $f_\phi$ function, which outputs the logit score for all the choices and the score with the highest choice is predicted as final output.

\subsection{\tb{Additional Results}}
Below we present the reconstructed samples (\figref{daren_ori_recon}) using \daren for different data sets that are representative of the median reconstruction error.

\begin{figure*}[!htp]
    \begin{center}
        \includegraphics[width=0.86\linewidth]{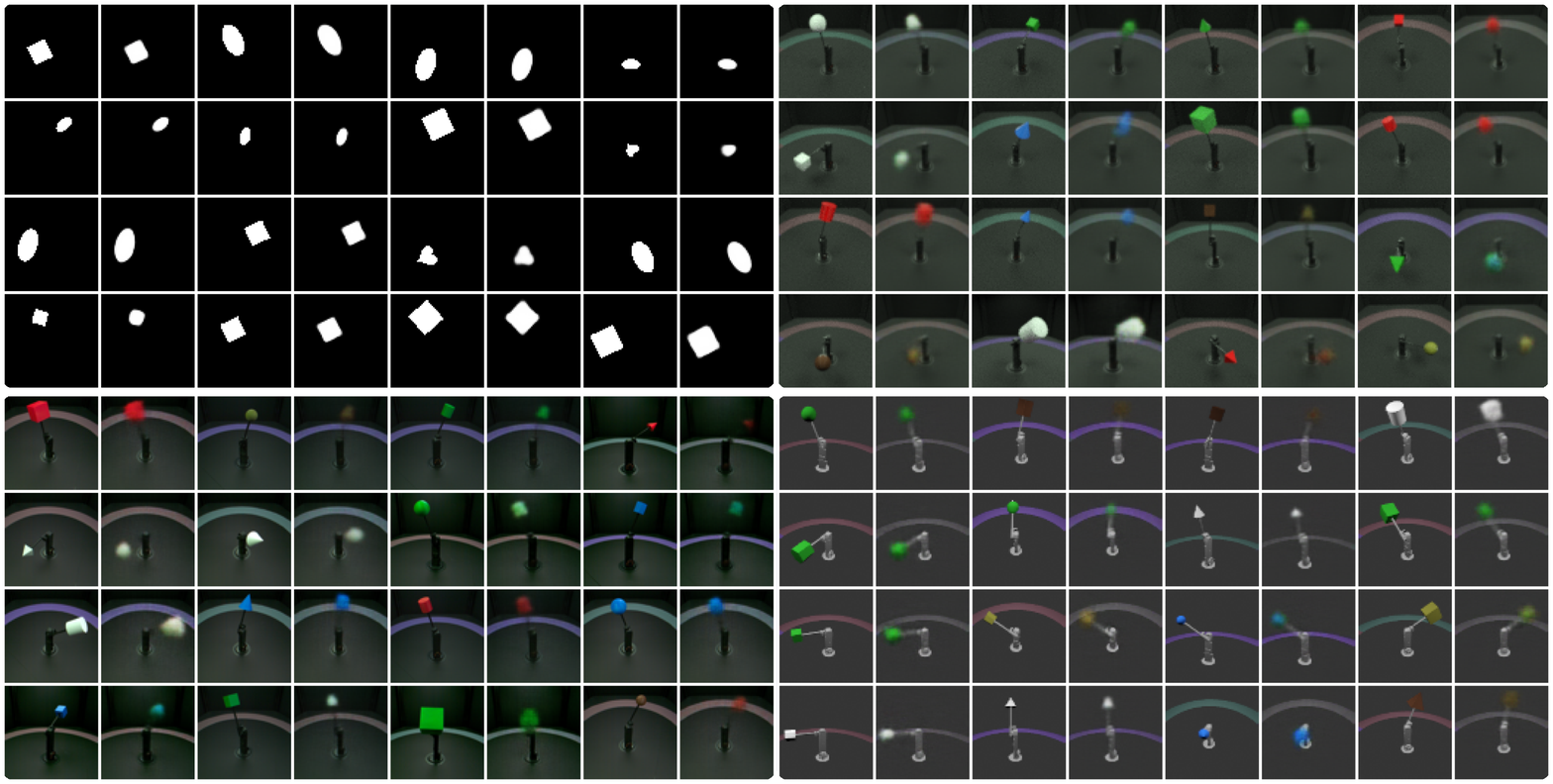}
    \end{center}
    \caption{Ground truth and reconstructions on the remaining four datasets: \textit{dsprites}, \textit{MPI3D-Realistic}, \textit{MPI3D-Real}, \textit{MPI3D-Toy} (Left to Right, Top to Bottom), using trained \daren at $300K$. Odd columns show real samples and even columns their reconstruction.}
    \label{fig:daren_ori_recon}
\end{figure*}

Next, we present the latent traversal of \daren over all the datasets, where the highlighted latent dimensions mark the presence of a generative factor our model successfully extracted, \figref{all_latent_travesal}.
\begin{figure*}[!hbp]
    \begin{center}
        \includegraphics[width=0.86\linewidth]{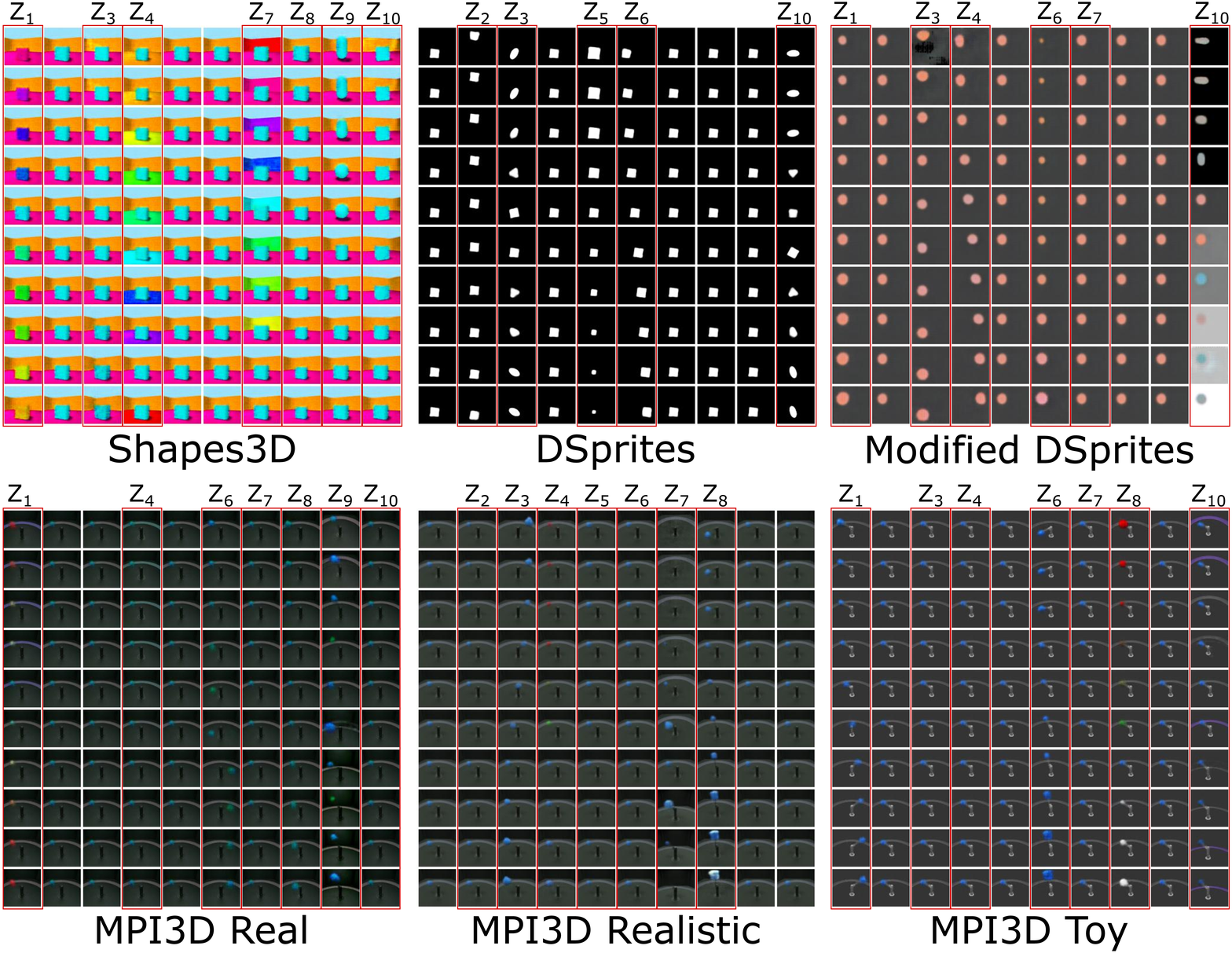}
    \end{center}
    \caption{Latent space traversal of \daren on all datasets. The visibily evident latent dimensions of variability are highlighted within color boxes where each exactly matches one of the four ground-truth factors. Shapes3D: $( Z_1, Z_3, Z_4, Z_7, Z_8, Z_9, Z_{10} )$, DSprites: $(Z_1, Z_3, Z_4, Z_7, Z_8, Z_9, Z_{10} )$, Modified DSprites: $( Z_1, Z_3, Z_4, Z_6, Z_7, Z_{10} )$, MPI3D Real: $( Z_1, Z_4, Z_6, Z_7, Z_8, Z_9, Z_{10}, )$, MPI3D Realistic: $( Z_2, Z_3, Z_4, Z_5, Z_6, Z_7, Z_8)$, MPI3D Toy: $(Z_1, Z_3, Z_4, Z_6, Z_7, Z_8, Z_{10})$}
    \label{fig:all_latent_travesal}
\end{figure*}

\figref{KL} displays the expected KL divergence for individual dimensions over all datasets. The left plot refers to the best performing model among all the models trained and the right plot presents the bad model trained. The count of bars in each plot (in best model) is same to the count of generative factors in each dataset which suggests our model minimizes duplicating or entangling more than factors. 
\begin{figure*}[!htp]
    \begin{center}
        \includegraphics[width=0.9\linewidth]{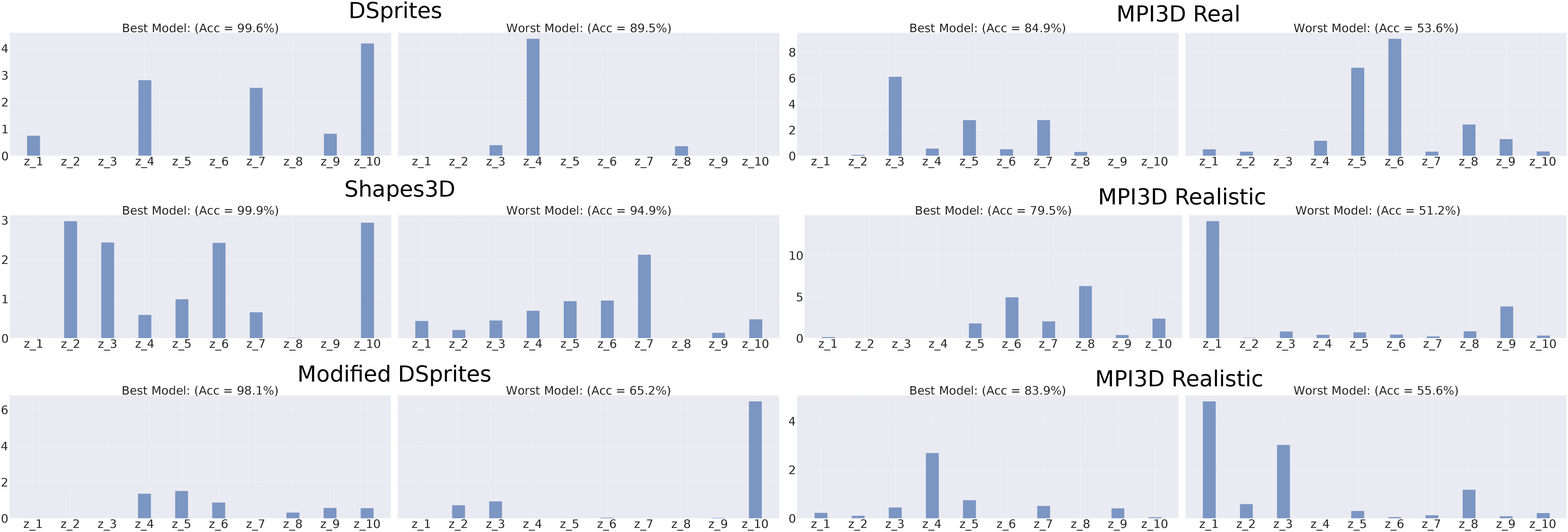}
    \end{center}
    \caption{Expected prior KL divergences for individual dimensions over all datasets. For each dataset, left plot represents the best performing model and right plot represents the lowest performing model over our hyper parameter sweep.}
    \label{fig:KL}
\end{figure*}
Finally, \figref{per_factor} displays the distribution of performance for each generative factor in the visual reasoning tasks.
\begin{figure*}[!hbp]
    \begin{center}
        \includegraphics[width=0.9\linewidth]{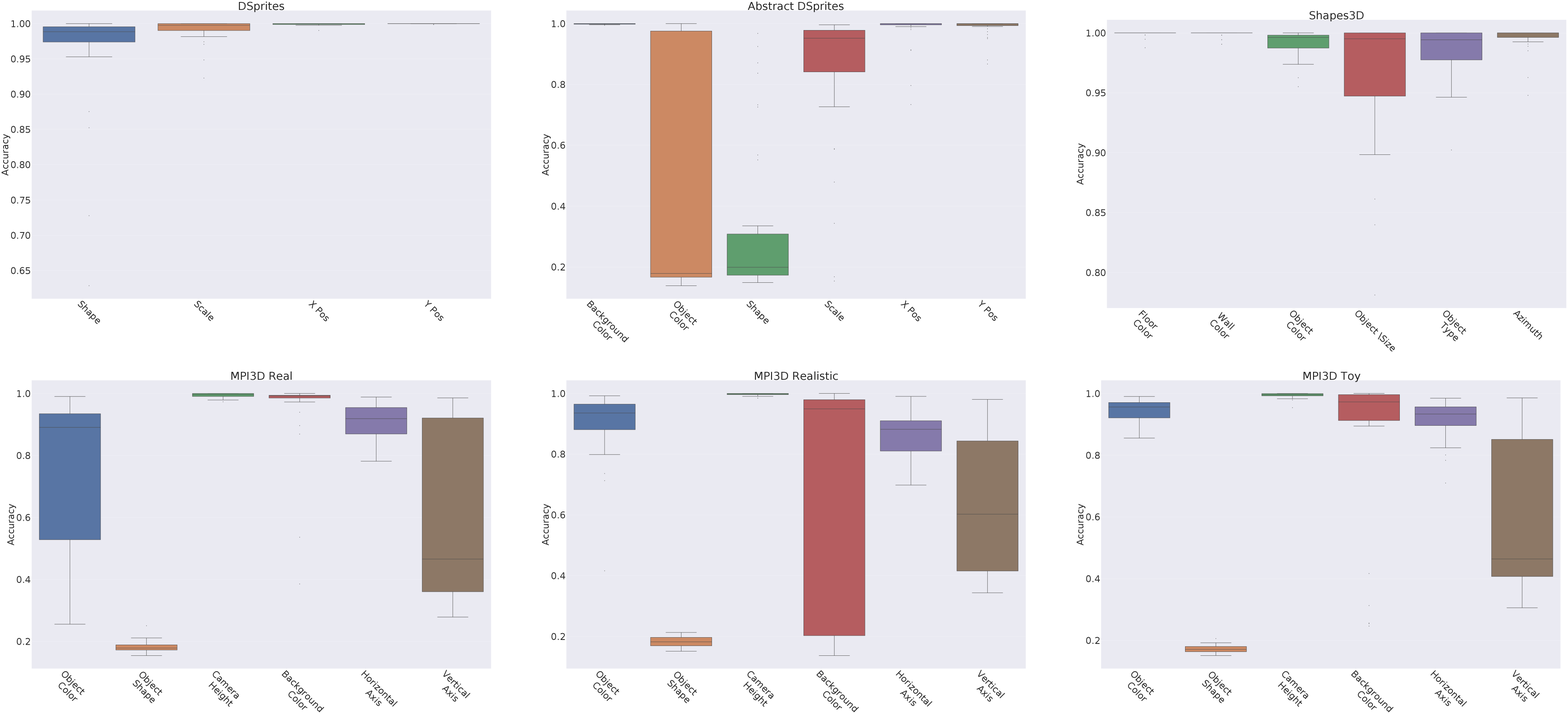}
    \end{center}
    \caption{Distribution of reasoning performance per generative attribute over all models for all datasets. It appears the discrete factor shape is the main reason behind drop in performance.}
    \label{fig:per_factor}
\end{figure*}


\fi

\bibliographystyle{IEEEtran}
\bibliography{icpr}

\appendix




%



\end{document}